\documentclass[10pt,twocolumn,letterpaper]{article}

\usepackage{wacv}              

\usepackage[accsupp]{axessibility}
\usepackage{graphicx}
\usepackage{amsmath}
\usepackage{amssymb}
\usepackage{booktabs}
\usepackage{multirow}
\usepackage{titling}
\usepackage[pagebackref,breaklinks,colorlinks]{hyperref}

\usepackage[capitalize]{cleveref}
\crefname{section}{Sec.}{Secs.}
\Crefname{section}{Section}{Sections}
\Crefname{table}{Table}{Tables}
\crefname{table}{Tab.}{Tabs.}

\def\wacvPaperID{2328} 
\def\confName{WACV}
\def\confYear{2024}

\begin{document}

\title{FastCLIPstyler: Optimisation-free Text-based Image Style Transfer Using Style Representations}

\author{Ananda Padhmanabhan Suresh$^*$
\and Sanjana Jain$^*$
\and Pavit Noinongyao
\and Ankush Ganguly
\and Ukrit Watchareeruetai
\and Aubin Samacoits
\\
\and
Sertis Vision Lab\\
597/5 Sukhumvit Road, Watthana, Bangkok, 10110, Thailand\\
{\tt\small \{asure,sjain,ponio,agang,uwatc,asama\}@sertiscorp.com}\\
{\small $^*$Both authors contributed equally to this work}
}
\date{}
\maketitle

\begin{abstract}
In recent years, language-driven artistic style transfer has emerged as a new type of style transfer technique, eliminating the need for a reference style image by using natural language descriptions of the style. 
The first model to achieve this, called CLIPstyler, has demonstrated impressive stylisation results. 
However, its lengthy optimisation procedure at runtime for each query limits its suitability for many practical applications. 
In this work, we present FastCLIPstyler, a generalised text-based image style transfer model capable of stylising images in a single forward pass for arbitrary text inputs. 
Furthermore, we introduce EdgeCLIPstyler, a lightweight model designed for compatibility with resource-constrained devices.
Through quantitative and qualitative comparisons with state-of-the-art approaches, we demonstrate that our models achieve superior stylisation quality based on measurable metrics while offering significantly improved runtime efficiency, particularly on edge devices.
\end{abstract}

\begin{figure*}[ht]
    \centering
    \includegraphics[width=0.95\textwidth]{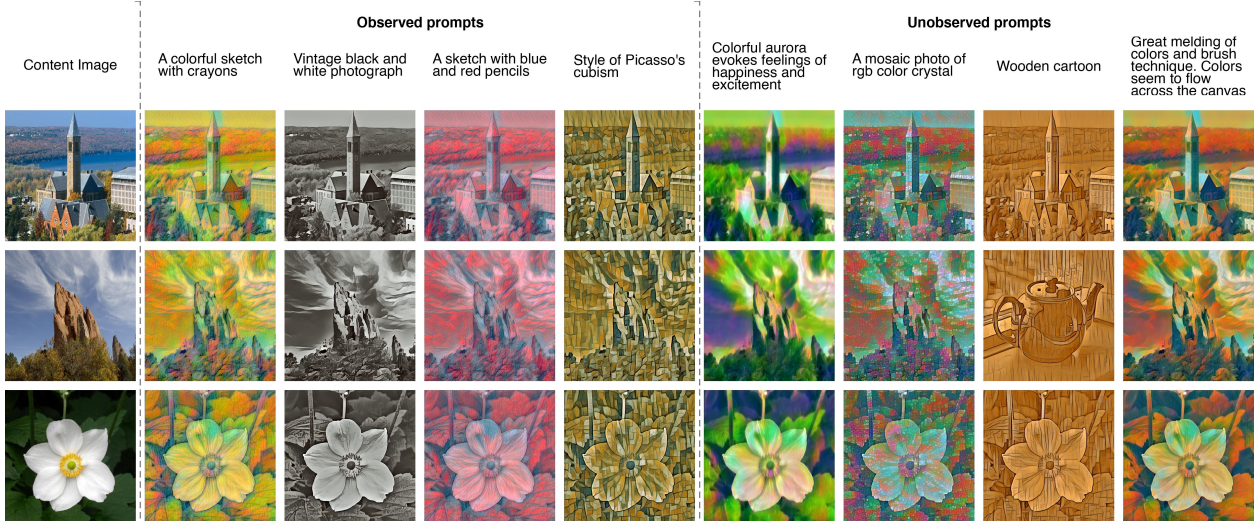}
    \caption{General performance of our model on observed and unobserved prompts. It is correctly able to identify styles across colours, art genres, and generic styles.}
    \label{fig:general_performance}
\end{figure*}

\section{Introduction}
\label{sec:intro}

The objective of style transfer is to recompose a content image with the semantic texture of a style image.
Research in this domain has been inspired by the work of Gatys \etal \cite{gatys_neural}, who demonstrated the capability of Convolutional Neural Networks (CNNs) to generate stylised images by extracting content information from arbitrary images and style information from well-known artworks \cite{jing_style_review}.
Their algorithm employs a pre-trained VGG-19 network \cite{simonyan2014very} to define content and style loss and jointly optimise them to create stylised images.
Since then, Li \etal \cite{li2017demystifying} reformulated the problem as distribution alignment, introducing new loss functions for the same.
Ulyanov \etal \cite{ulyanov2016texture} and Li \etal \cite{li2016precomputed} presented models that can apply a reference image's style in a single neural network pass, while Ghiasi \etal \cite{ghiasi_google}, Huang \etal \cite{huang2017arbitrary} (AdaIN) and Chen \etal \cite{chen2016fast} developed models that can transfer style from an arbitrary style image without requiring optimisation at runtime. 
More recent models, such as SANet  \cite{park2019arbitrary} and AdaAttn \cite{liu2021adaattn}, have implemented attention mechanisms to improve the quality of synthesised images. 
While these approaches successfully create visually pleasing stylised images, they rely on the availability of the desired reference style image, which is not always available. 

Recognising this limitation, \cite{kwon2021clipstyler} developed a method called CLIPstyler to solve Language Driven Artistic Style Transfer (LDAST) \cite{fu2022language}, the concept of which is to stylise images based on a text prompt instead of a reference style image.
CLIPstyler employs CLIP \cite{radford2021learning}, an embedding model that projects image and text to a shared embedding space, enabling the application of a style text prompt to a content image.
One drawback of this is its time-consuming optimisation procedure at inference time for each text query, making it unsuitable for real-world applications.
To address this issue, Fu et al. \cite{fu2022language} recently introduced Contrastive Language Visual Artist (CLVA), capable of stylising images using a general text prompt without optimisation.
However, their model does not generalise well to unseen prompts and is not able to support resource-constrained devices, losing out on a lot of practical applications for LDAST.

\begin{figure}[t]
    \centering
    \includegraphics[width=0.925\linewidth]{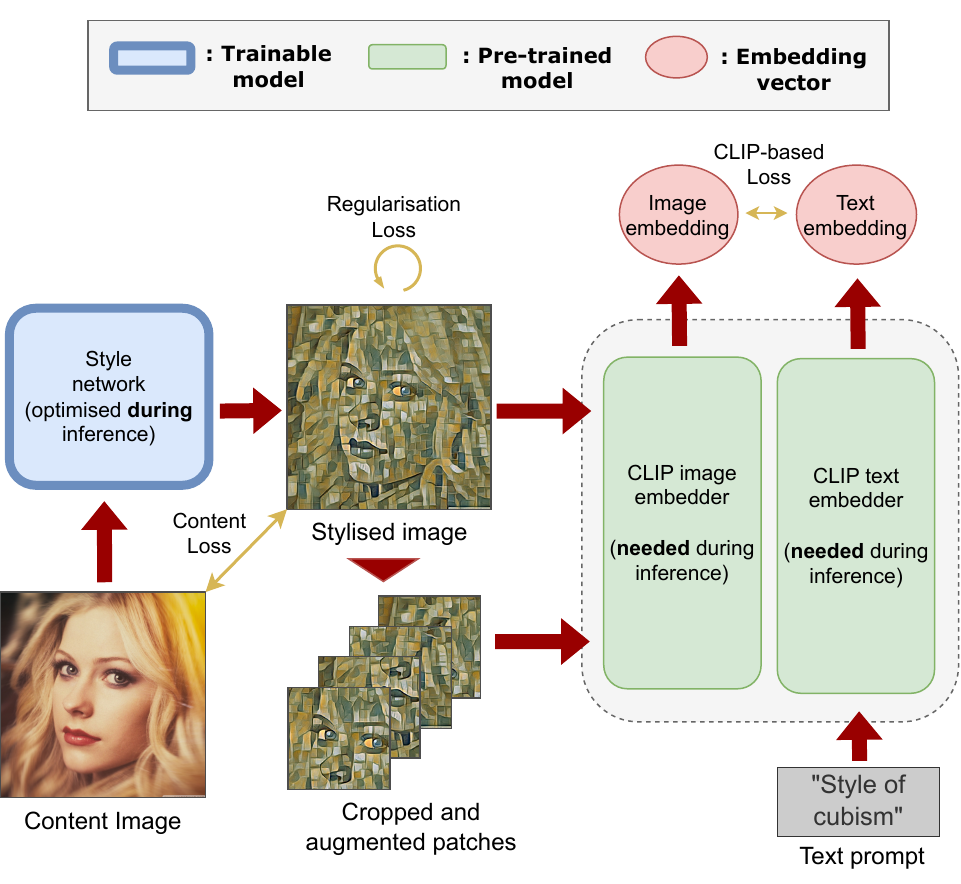}
    \caption{The architecture diagram of CLIPstyler model.}
    \label{fig:clipstyler}
\end{figure}

\begin{figure}[t]
    \centering
    \includegraphics[width=0.95\linewidth]{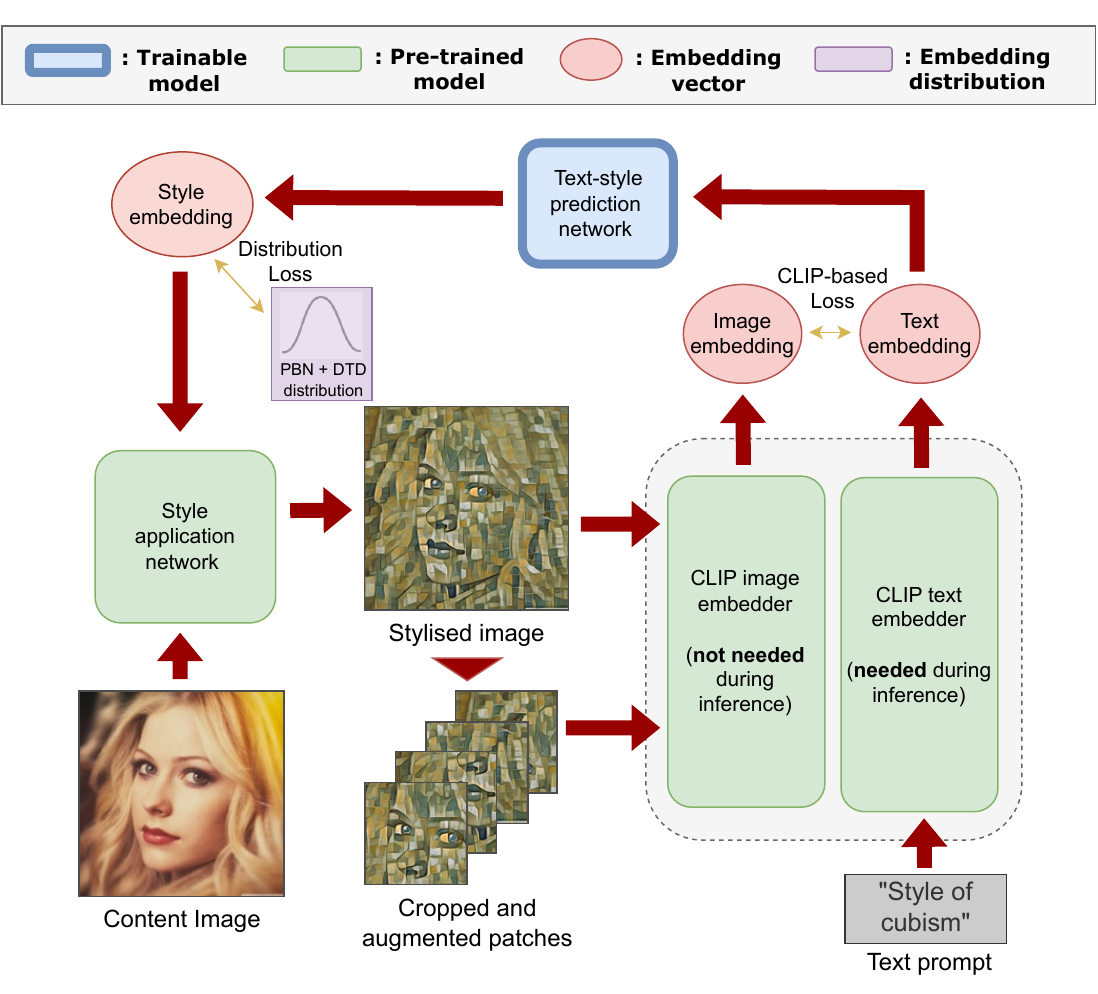}
    \caption{The architecture diagram of our FastCLIPstyler model.}
    \label{fig:our_model}
\end{figure}

In this work, we introduce FastCLIPstyler, a text-based image style transfer model which, unlike CLIPstyler, eliminates the need for runtime optimisation.
We achieve this by incorporating a pre-trained, generalisable, purely-vision based style transfer network into the CLIPstyler framework. 
The CLIP model facilitates this approach, enabling training without dependence on labelled datasets such as Artemis \cite{artemis} and Describable Textures Dataset (DTD) \cite{dtd}, which usually requires a time-consuming and expensive annotation procedure to create.
We also outline a strategy for generating a text prompt dataset, which begins with a set of basic prompts and combines them into more intricate combinations.
In addition, we also demonstrate that our model has superior generalisation ability to unseen prompts, especially when compared to the current state-of-the-art in LDAST, CLVA.
Furthermore, to broaden our impact, we also introduce EdgeCLIPstyler: a streamlined, edge-compatible iteration of FastCLIPstyler, aptly designed for low-powered devices with minor architectural tweaks.
As illustrated in \cref{fig:general_performance}, our approach adeptly applies styles from both familiar and unfamiliar prompts, yielding visually pleasing results.

We highlight the following as the main contributions of our paper: 
\begin{itemize}
    \item We introduce FastCLIPstyler, a text-based image style transfer model that can stylise images in a single forward pass through a network for arbitrary text inputs, without the need for runtime optimisation.

    \item In our approach to the LDAST problem, we formulate FastCLIPstyler so that it only requires a dataset of text prompts for training, eliminating the need for corresponding style images. Instead, we employ a novel label generation process, optimising the text-style prediction network for each datapoint individually.

    \item Using both qualitative and quantitative evaluations, we demonstrate that FastCLIPstyler outperforms CLIPstyler in terms of image generation speed at runtime (730x faster) and surpasses CLVA \cite{fu2022language} on stylised image quality. 
    
    \item We also introduce an edge-compatible variant, EdgeCLIPstyler, which is, to the best of our knowledge, the first model to enable LDAST on edge devices.
\end{itemize}
\label{section:related}

\section{Method}
\label{section:method}

\subsection{Background knowledge}
\label{section:background}

In this section, we outline the key components our model adopts from established works: Ghiasi's style transfer network \cite{ghiasi_google} CLIPstyler \cite{kwon2021clipstyler}

\textbf{Ghiasi's style transfer network:} Ghiasi \etal \cite{ghiasi_google} introduced one of the first purely vision-based style transfer networks capable of transferring the style of any reference image to a content image in a single forward pass.
The model consists of a style prediction network that converts a reference style image into a 100-dimensional style embedding and a style application network that applies the embedding to the content image.
As Ghiasi \etal's style prediction network relies on a style image, it is not applicable to us; we just adopt their style application network, which takes the content image and a style embedding as inputs.

\textbf{CLIPstyler:} While most style transfer approaches utilise networks like VGG to model their transfer process, CLIPstyler \cite{kwon2021clipstyler} employs the CLIP model \cite{radford2021learning} for style transfer, enabling the use of natural language descriptions instead of reference images. 
The content image is processed through an image-to-image CNN that directly transforms it into a stylised version. 
CLIP computes a similarity score between the generated image and the user's text query, and the CNN is optimised until the generated image resembles the textual description. 
The architecture of CLIPstyler is shown in \cref{fig:clipstyler}.

\subsection{FastCLIPstyler}
\label{subsection:architecture}

The CLIPstyler framework trains a CNN model from scratch during inference to transform content images into stylised versions. 
However, leveraging existing style transfer models can eliminate this ground-up training. 
For instance, the Ghiasi et al. model \cite{ghiasi_google} uses a style application network that stylises content images based on a 100-dimensional style representation. 
Instead of training a full CNN model, we train a compact, fully-connected feed-forward network called the `text-style prediction network', which inputs CLIP text embeddings and predicts their 100-dimensional style representations. 
The FastCLIPstyler architecture is shown in \cref{fig:our_model}.
We selected Ghiasi's network for its robust generalisation capabilities, demonstrated by its performance on a variety of unseen style images. 
Our pipeline can be easily adapted to other image-based style transfer methods, provided they offer an explicit style representation.

Our text-style prediction network can be trained to learn the mapping between text embeddings and their corresponding style representation in the Ghiasi-style space.
For doing this, the network can be trained using a dataset of text prompts and their corresponding style embeddings.
In order to get these style embeddings, we optimise the untrained model to each of the prompts, one at a time, with the guidance of CLIP.
Optimising the neural network for each query in this manner is similar to how it was done in CLIPstyler, which served as the inspiration for this approach.
While in CLIPstyler, the optimisation procedure for a particular query generated the corresponding final image, what we want from our network at this stage is the intermediate style embedding corresponding to a query.
This creates `labels' for each query in the dataset, which can then be used to train a generalised network.
Compared to CLIPstyler, our trainable network has a much more simplified task to learn --- it only needs to predict an appropriate input for the pre-trained style application model without worrying about preserving content or artistic quality in the final image.

The text-style prediction network that we use is a simple, fully-connected feed-forward network comprising four hidden layers that takes a 512-dimensional text embedding as input and outputs a 100-dimensional style representation. 
The activation function in between the layers is the Leaky ReLU \cite{maas2013rectifier} with a negative slope of 0.2.
The final layer uses the tanh activation function to normalise the style representations between -1 and 1.
The full feed-forward model specification, along with the hyperparameters, can be found in the supplementary section. 

In \cref{fig:clipstyler} and \cref{fig:our_model}, the architectural differences between CLIPstyler and our proposed FastCLIPstyler are highlighted.
FastCLIPstyler needs one-time training using a dataset and then requires no inference-time optimisation, while CLIPstyler optimises a CNN for each input during inference, making it more time-consuming. 
This is because CLIPstyler transforms content images into stylised ones by optimising a dedicated network each time, whereas FastCLIPstyler converts an embedding in the CLIP space to the style application network, a less complex task. 
This results in FastCLIPstyler being more efficient at inference.

\subsection{Dataset generation}
\label{dataset}

In training our model, we capitalise on a key advantage: our method does not require a labelled image dataset, as it can generate its own labels by leveraging CLIP's capabilities.
In this regard, CLIP essentially acts as our `labeller'. 
The initial dataset we need consists of a diverse set of style prompts, which can be easily generated from a corpus, as opposed to images that must be collected from the real world.

To assemble this list of prompts, we combine keywords encompassing colours, textures, art styles, and objects with distinct textures, ultimately generating a dataset of 4,300 prompts. 
For example, by pairing art styles and real-world objects, we create prompts like `mosaic stone wall' or `acrylic snow'.
To further enhance the generalisation and style transfer quality of our model, we employ ChatGPT's \cite{brown2020language} language generation capabilities, which allows us to create an additional 1,500 style prompts and increase the dataset's diversity. 
These prompts cover aspects of nature, emotional states, and examples similar to those found in the ArtEmis dataset.
We demonstrate that this relatively simple strategy of data generation is adequate to train the model so that it's capable of generalising across a wide range of seen and unseen prompts. 

To the best of our knowledge, our approach is unique in that it does not depend directly on labelled datasets like ArtEmis or DTD. 
The ability to self-generate a dataset for training an LDAST model represents a significant advantage, as it eliminates the need for a costly data collection phase. 
Additional information on our data generation strategy can be found in the supplementary section.

\subsection{Loss function}
\label{subsection:loss}

In this section, we begin by exploring the loss function we use to optimise the model for each text prompt to generate their corresponding style embedding labels. 
We later discuss the loss function used to train the generalised model using these style embedding labels.

As proposed by Kwon and Ye \cite{kwon2021clipstyler} and originally suggested by StyleGAN-NADA \cite{gal2021stylegan}, we adopt the directional loss $L_{dir}$ to measure the closeness of the image generated by the style transfer model with the text input using CLIP in a stable way.

\begin{figure*}[t]
    \centering
    \includegraphics[width=0.9\linewidth]{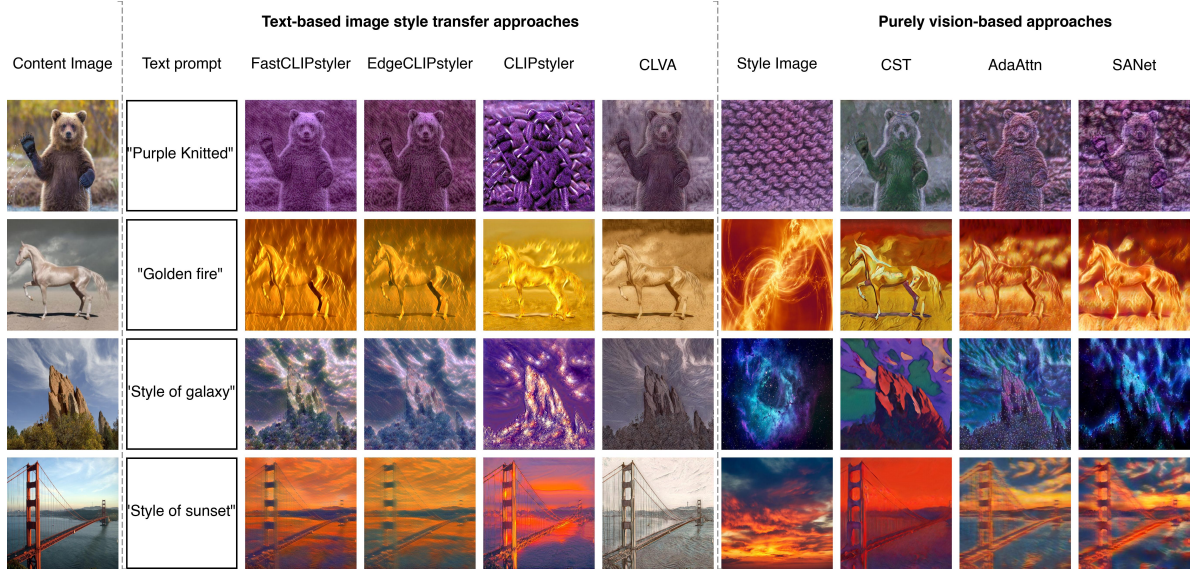}
    \caption{Comparison of our model with other state-of-the-art models in its ability to transfer styles of various complex queries and artworks.}
    \label{fig:general_comparison}
\end{figure*}

We also adopt PatchCLIP loss, $L_{patch}$, a novel loss term proposed by  \cite{kwon2021clipstyler} that was shown to greatly improve the stylisation results.
CLIPstyler also defines a content loss that is meant to ensure that the content is preserved during the transfer and a total variation regularisation loss to alleviate unwanted artefacts. 
However, we find that these loss terms are unnecessary for our training since the Ghiasi network has already been trained to preserve content, as well as to apply the style in a manner that is visually pleasing.
Hence we only use the CLIP-based losses, $L_{dir}$ and $L_{patch}$, from \cite{kwon2021clipstyler}.

However, using these two loss functions alone poses a practical challenge.
The entire 100-dimensional space of real numbers does not constitute a valid input into the Ghiasi network.
The samples must be drawn from a specific region of the $\mathbb{R}^{100}$ for the model to work as intended.
However, there is nothing stopping the text-style prediction network from predicting style embeddings that are well out of the valid input region of the Ghiasi network.
To address this, we construct another loss term that penalises the network for making predictions outside the valid input region of the Ghiasi network.
Since the model was trained on the PBN \cite{pbn} and DTD \cite{dtd} datasets, we compute the style embeddings of all images in these datasets and assume a normal distribution over these in order to calculate the likelihood of a predicted embedding vector being sampled from this distribution.
This likelihood defines a distribution loss term  $L_{dis}$:
\begin{equation}
    L_{dis} = (x-\mu_{data})^T \Sigma_{data}^{-1} (x-\mu_{data}),
\end{equation}
where $x$ is the embedding tensor, $\mu_{data}$ and $\Sigma_{data}$ are the mean and covariance of the embeddings of the PBN and DTD datasets.
This term penalises the model for predicting style embeddings that lie far from the distribution of style embeddings of real images. 

Putting these together, our overall loss function is formulated as:
\begin{equation}
    \label{eq:loss_total}
    L_{total} = \lambda_{dir} L_{dir} + \lambda_{patch} L_{patch} + \lambda_{dis} L_{dis},
\end{equation}
where $\lambda_{dir}$, $\lambda_{patch}$ and $\lambda_{dis}$ are coefficients governing the weights of their losses respectively.

To generate text embeddings, we pass the text prompts through the CLIP text embedder.
Once we have the dataset of text prompts and their corresponding text and style embeddings, we train the generalised text-style prediction network using a simple mean squared average error loss.

\begin{figure*}[t]
    \centering
    \includegraphics[width=0.9\linewidth]{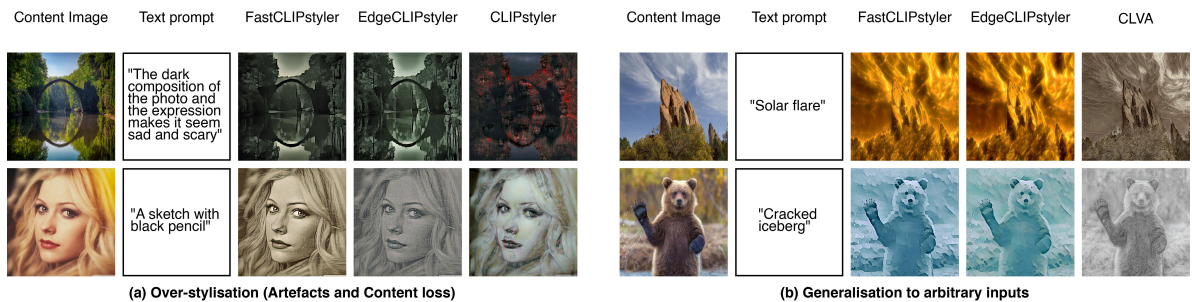}
    \caption{Examples where limitations of CLIPstyler and LDAST are highlighted and compared to our model.}
    \label{fig:clip_clva_bad}
\end{figure*}

\subsection{EdgeCLIPstyler: Edge-compatible FastCLIPstyler}

FastCLIPstyler, built upon the powerful CLIP model, yields impressive stylisation results but is computationally expensive, making it unsuitable for resource-constrained edge devices. 
To address this limitation, we introduce EdgeCLIPstyler, an edge-compatible FastCLIPstyler framework adapted for edge devices using the more resource-efficient Sentence-BERT paraphrase-albert-v2 \cite{reimers-2019-sentence-bert} text-embedder model.

Compared to CLIP's text embedder, the Sentence-BERT paraphrase-albert-v2 embedder is smaller and demands fewer computational resources, making it ideal for our edge-based approach. 
We train the text-style prediction network using embeddings generated from the Sentence-BERT model.
Although the CLIP model is still needed for supervising model training (in the CLIP loss computation), any text embedder can be utilised during model inference. 
This slight architectural modification enables us to achieve an edge-compatible LDAST model.

As we demonstrate in \cref{section:results}, EdgeCLIPstyler is capable of running on edge devices while delivering impressive stylisation results; it can be an excellent tool for social media and graphic designing applications.
This also finds applications in tools like video conferencing background enhancement, social media filters, photo editing applications etc., directly on mobile devices, without relying on remote servers for processing, which has implications for privacy and data security.

\section{Results}
\label{section:results}

All experiments were performed on a 6-core Intel i5 desktop with 16 GB RAM and a single 8GB Nvidia GeForce RTX 2070 GPU. 
For benchmarking our inference time on low-powered CPU-only devices, we use the Intel NUC Kit NUC7i5BNH with 8 GB RAM and Raspberry Pi 3B+ with 1 GB RAM.

During the dataset generation step, for the CLIP image embedder, we chose the ViT-B/32 backbone to generate the 512-dimensional style embeddings, making our experiments comparable to the CLIPstyler framework.
Our choice for the CLIP model is based on the trade-off between computational efficiency and model performance presented by \cite{radford2021learning}.

\subsection{Qualitative evaluation}
\label{subsection:qualitative}

In \cref{fig:general_performance}, we show the style transfer capability of our model.
It has an awareness of a wide range of queries and is able to apply colours, textures, art styles, and real-world objects, as well as compound statements combining these.
We also show that the model has good performance when it comes to handling prompts that it has never seen during training. 
\Cref{fig:general_comparison} shows the comparison of our approaches with other state-of-the-art techniques in style transfer.
We also compare our models with recent examples of purely vision-based style transfer techniques, namely, CST \cite{svoboda2020two}, SANet \cite{park2019arbitrary}, and AdaAttn \cite{liu2021adaattn}.
Despite not having the advantage of an explicit reference style image during style transfer, our model generates images similar to these techniques.

\Cref{fig:clip_clva_bad} demonstrates the superior performance of our models in certain instances compared to CLIPstyler and CLVA. 
Specifically, in \Cref{fig:clip_clva_bad}(a), CLIPstyler, influenced by the first prompt, introduces disturbing faces, and the second prompt leads to further undesirable artefacts. 
In these cases, our models maintain image integrity without such issues.

\Cref{fig:clip_clva_bad}(b) emphasises instances where CLVA's style application fails to align with the given prompt.
We observe that CLVA particularly struggles when the prompt significantly deviates from the ArtEmis and DTD datasets, which served as its training source. 
In contrast, as depicted in \cref{fig:general_performance}, our model demonstrates the capacity to adapt to new, open-ended prompts that are reminiscent of the ones from the ArtEmis dataset, such as `Colorful aurora evokes feelings of happiness and excitement.'

\begin{figure}[t]
    \centering
    \includegraphics[width=0.95\linewidth]{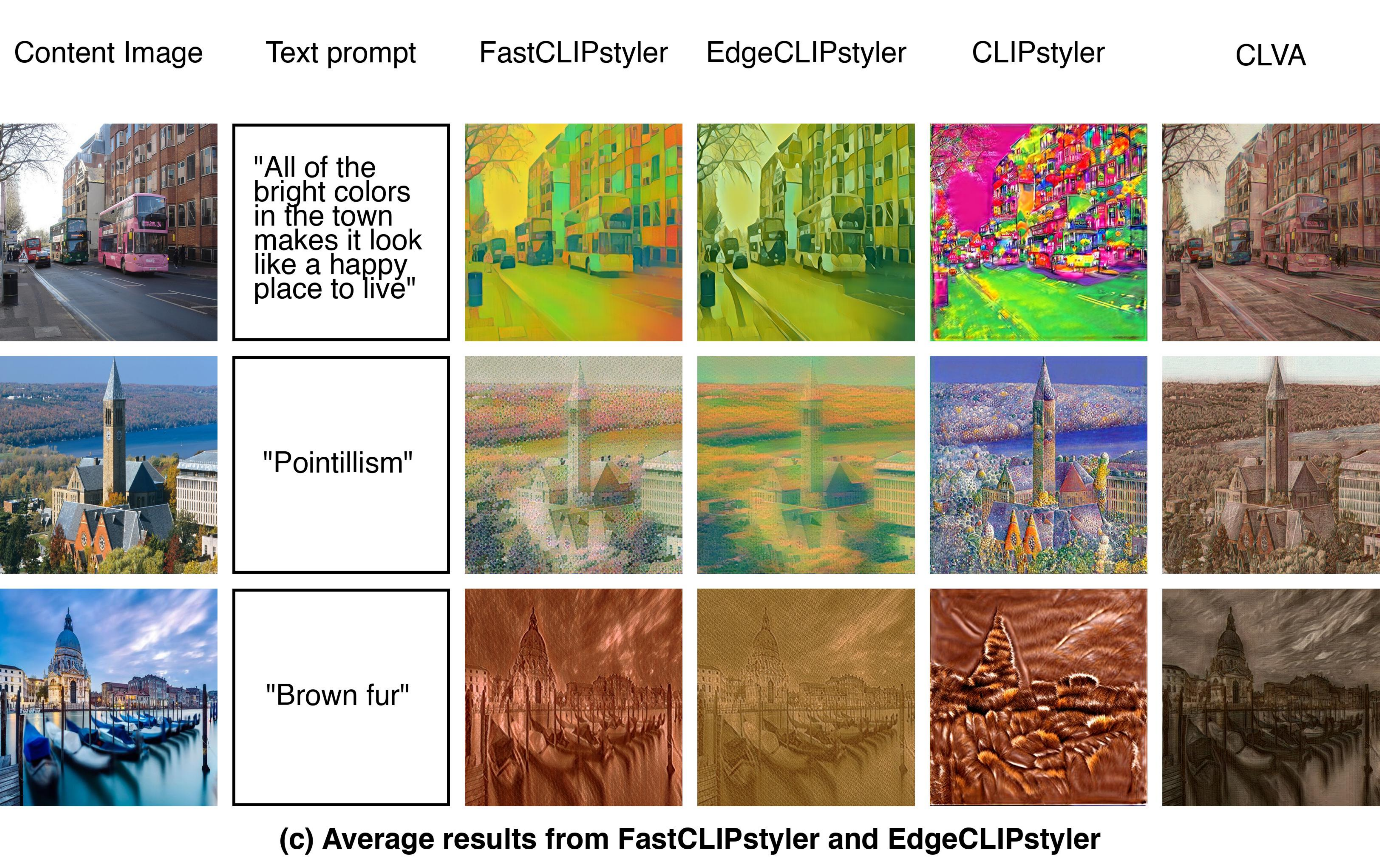}
    \caption{Examples where some of the limitations of our models are highlighted.}
    \label{fig:fclip_doing_bad}
\end{figure}

In \cref{fig:fclip_doing_bad}, we acknowledge some limitations of our model compared to CLIPstyler and CLVA. 
Our models tend to produce more monochromatic images when a specific colour is implied directly or indirectly in the prompt, as seen in the `brown fur' and `wooden cartoon' examples.
EdgeCLIPstyler occasionally falls behind FastCLIPstyler, as in the `pointillism' case. 
Additionally, CLVA's images exhibit a superior `artistic quality,' as demonstrated in the `bright colours in the town' prompt. 
However, CLVA fails to maintain consistency with the prompt when moving beyond the training distribution of ArtEmis and DTD, whereas our models remain more robust.

\subsection{Quantitative evaluation}
\label{subsection:quantitative}

\begin{table*}
\centering
\begin{tabular}{cccccc}
\hline 
Method & VLS $\uparrow$ & SSIM $\uparrow$ & Percept $\downarrow$ & FAD $\downarrow$ & Human evaluation $\uparrow$ \\
\hline
CLVA           & 0.092                & \underline{0.375}  & 24.93               &   \underline{1.032}     & 2.96     \\ 
CLIPstyler     & \textbf{0.194}       & 0.218              & 23.56               &   1.035                 & 3.17     \\ 
FastCLIPstyler & \underline{0.113}    & \textbf{0.418}     & \textbf{21.83}      &   1.033                 & \textbf{3.39}   \\ 
EdgeCLIPstyler & 0.091                & 0.368              & \underline{23.14}   & \textbf{1.024}          & \underline{3.18}       \\ 
\hline
\end{tabular}
\caption{Various metrics that we use to quantify the performance of different models}
\label{tab:quantitative_results}
\end{table*}

For the quantitative evaluation of our proposed text-based image style transfer model, we create a dataset containing 80 prompts, including general prompts and prompts inspired by the DTD \cite{cimpoi14describing} and ArtEmis \cite{artemis} datasets. 
In order to perform an automatic metric-based evaluation, we compute four key metrics: Vision Language Semantic (VLS), Structural SIMilarity (SSIM), Percept, and FAD. 
VLS \cite{vls} is the CLIP \cite{radford2021learning} cosine similarity between style instructions and the stylisation results. 
SSIM \cite{ssim} compares the image quality based on contrast, luminance, and structural aspects between two images. 
Percept \cite{fu2022language} computes the style reconstruction loss \cite{perceptual_justin} from the gram matrix of visual features between two images. 
Lastly, FAD \cite{fu2022language} computes the L2 distance between the activation maps of stylised images obtained from InceptionV3.
To compute the SSIM, Percept, and FAD metrics, we utilise a stylised image ground truth for comparison. 
Following \cite{fu2022language}, we generate semi-ground truth (Semi-GT) results using the state-of-the-art purely vision-based style transfer method, AdaAttn \cite{liu2021adaattn}, directly from style images. 
Due to the inherent limitations of the automatic metrics, we also conducted a user study for the quantitative evaluation through an online user survey performed with 75 participants.
Each participant was shown a set of 20 randomly selected prompts, each consisting of results from the four models: CLIP, CLVA, FastCLIPstyler, and EdgeCLIPstyler. 
They were asked to rate the quality of the stylised images on a scale of 1 to 5.
More details on how the quantitative experiments were set up can be found in the supplementary section.

In \cref{tab:quantitative_results}, we show the quantitative evaluation of our model in comparison to other baselines for the LDAST task. 
From the automatic metrics evaluation, our FastCLIPstyler excels in preserving structural similarity (highest SSIM) with the Semi-GT. CLIPstyler demonstrates the closest match to the text prompt based on the VLS metric, as it directly optimises for VLS for each datapoint.
At the same time, our model achieves the lowest stylisation Percept loss, indicating strong stylisation performance on par with the Semi-GT. 
CLIPstyler, however, experiences the highest Precept loss, likely indicating over-stylisation. 
EdgeCLIPstyler achieves the lowest overall similarity distance (indicated by FAD) with the Semi-GT, while FastCLIPstyler achieves a close and comparable FAD distance.
In terms of user study results, our methods surpass other techniques.
One contributing factor seems to be that CLVA struggles to perform well when faced with unseen prompts.
Another reason is that while CLIPstyler is a viable option, it occasionally generates stylised images containing random artefacts, leading to odd-looking results.

Although EdgeCLIPstyler is capable of achieving stylisation mostly on par with FastCLIPstyler, it is worth noting that EdgeCLIPstyler faces challenges when dealing with some specific queries. This is because the underlying BERT model lacks the robustness in capturing visual perception while associating text with images like a CLIP text embedder \cite{bertvclip}.

\begin{table*}
\centering
\begin{tabular}{crrrr}
\hline
Device       & FastCLIPstyler & EdgeCLIPstyler & CLVA & CLIPstyler     \\ \hline 
RTX 2070 SUPER   & 70.444  & 24.891  & 25.088 & $51.360 \times 10^{3}$\\ 
6X i5-9500 CPU  & 1770.740        & 390.328        & 5520.347 & $13.098 \times 10^{5}$\\ 
Intel NUC    & 7138.115           & 1297.359             & 20998.551   & $44.26 \times 10^{5}$  \\ 

Raspberry PI 3B+    & NA           & $15.011 \times 10^3$             & NA & NA \\ \hline 
\end{tabular}
\caption{Benchmarking for the time taken (in milliseconds) for various techniques.}
\label{tab:performance_on_edge}
\end{table*}

\begin{figure}[t]
    \centering
    \includegraphics[width=1\linewidth]{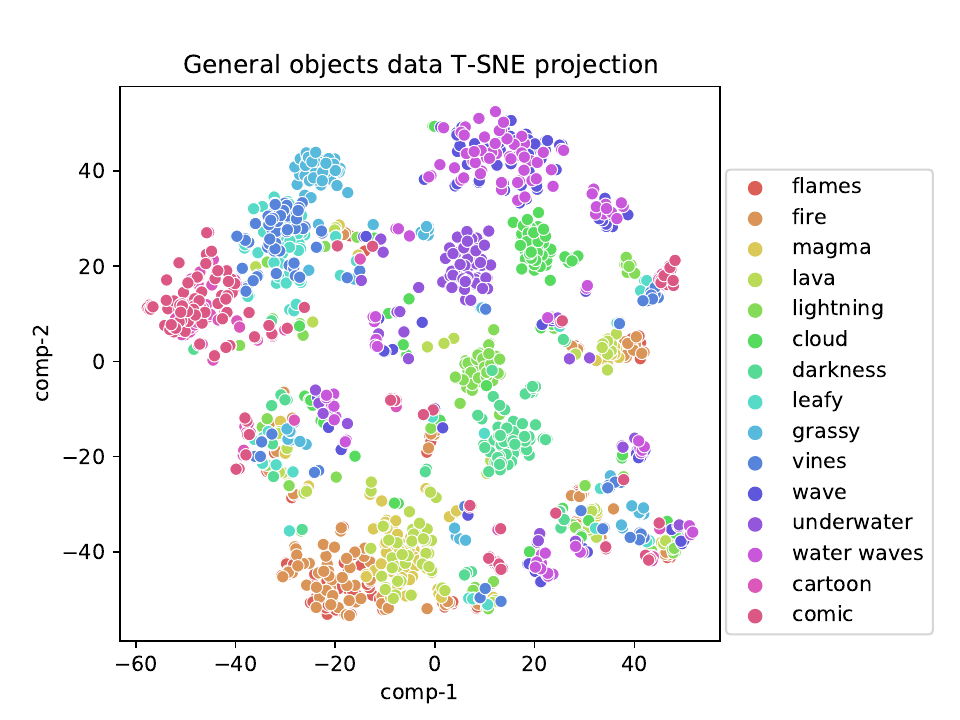}
    \caption{General objects t-SNE visualisation}
    \label{fig:t_sne_bmvc_general_prompts}
\end{figure}

\begin{figure*}
\centering
\begin{tabular}{cc}

\includegraphics[width=0.45\linewidth]{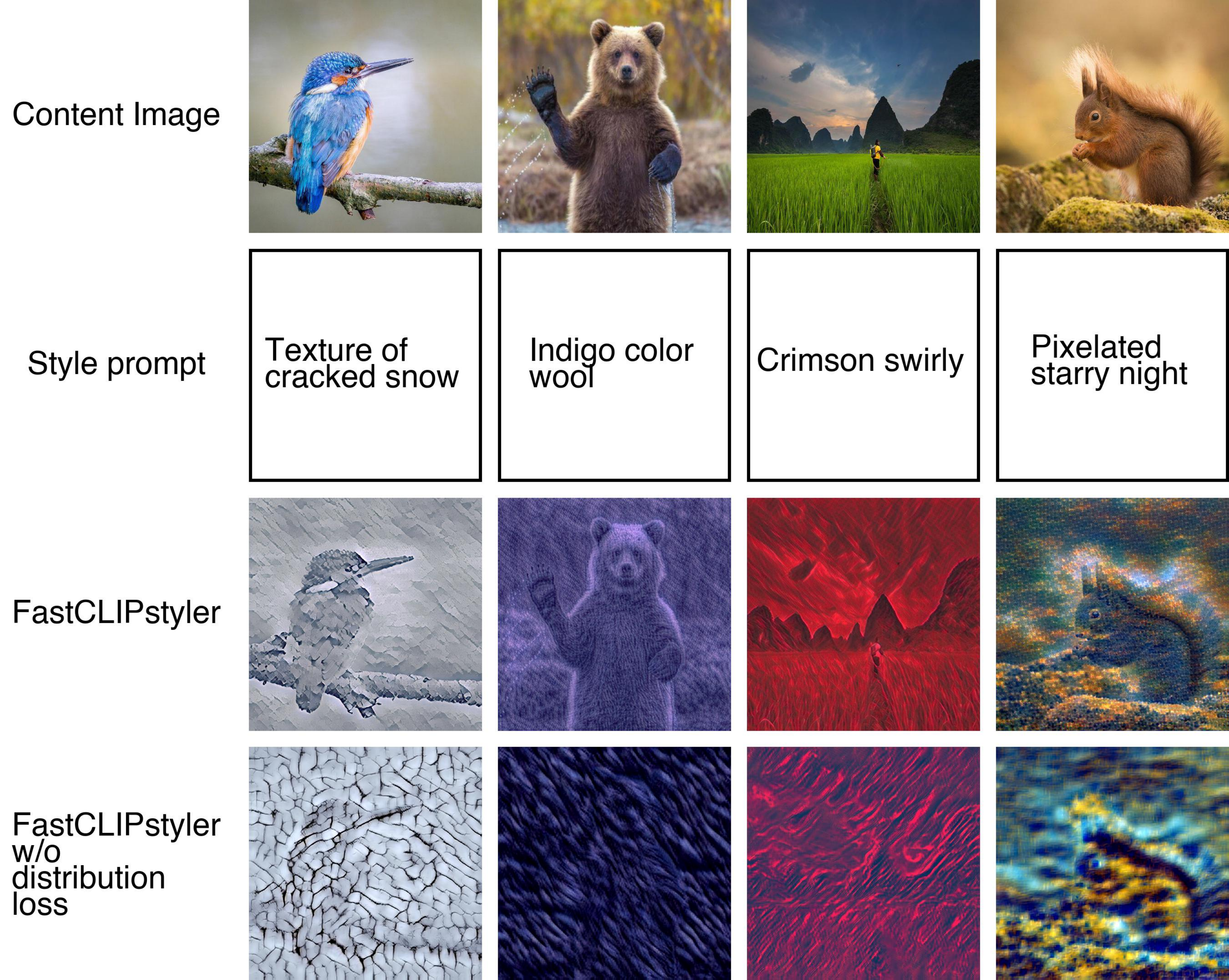}&
\includegraphics[width=0.45\linewidth]{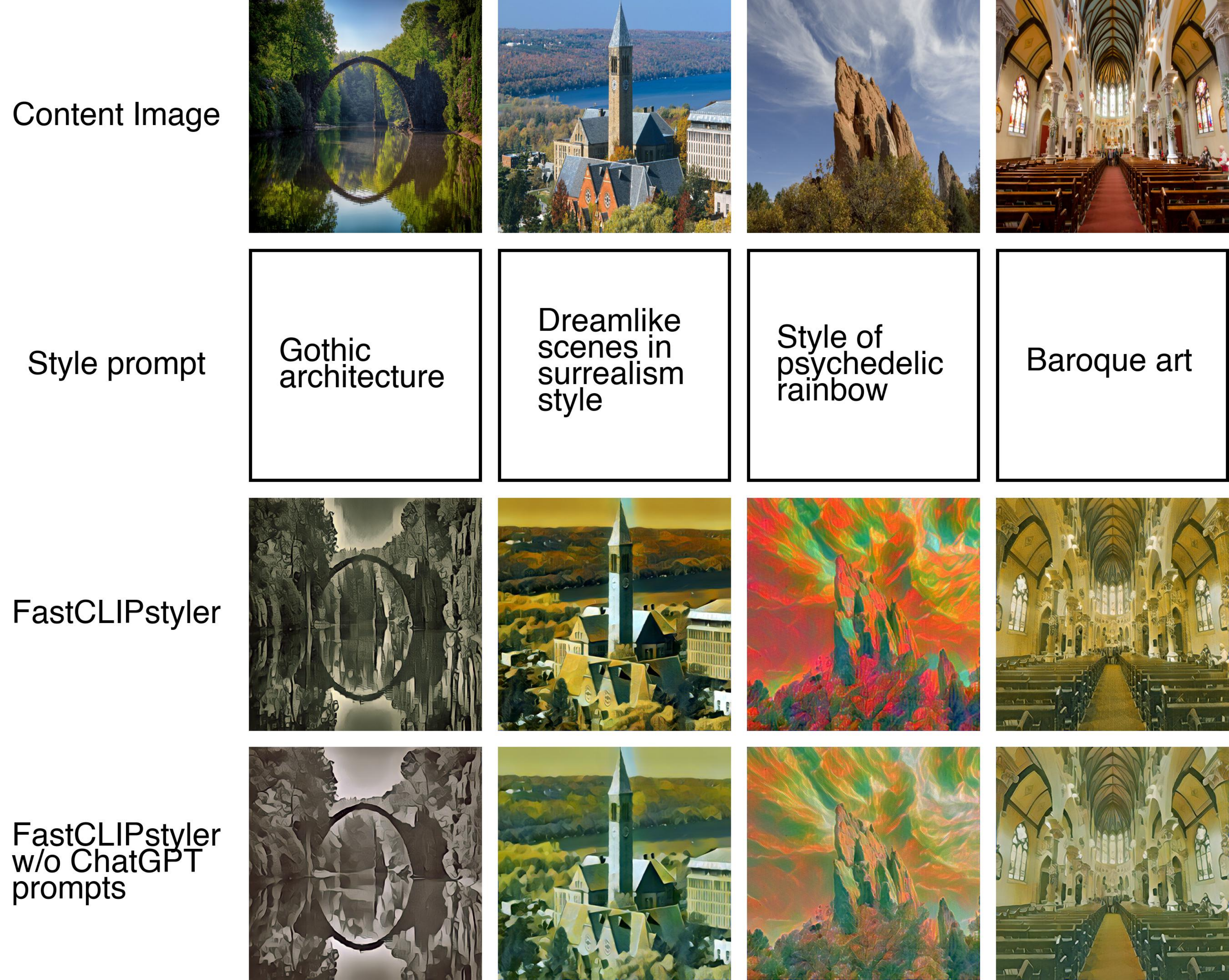}\\
(a) Distribution loss & (b) ChatGPT prompts for training \\
\\
\end{tabular}
\caption{Comparison of the stylised images with various content and text prompts. The results from the absence of crucial components in our pipelines are shown in different columns.}
\label{fig:ablation_study}
\end{figure*}

\subsection{Inference time benchmarking}

We perform inference time benchmarking for various models on different platforms, which can be found in \cref{tab:performance_on_edge}.
Our FastCLIPstyler, though slower than CLVA on the powerful RTX 2070 SUPER GPU, still outperforms CLIPstyler by 730x. 
Additionally, it is more adaptable to low-powered CPU-only devices, running three times faster than CLVA on the edge devices we tested on.
Our proposed EdgeCLIPstyler model, which uses the Sentence-BERT paraphrase-albert-v2 text embedder \cite{reimers-2019-sentence-bert} during inference, is edge-compatible and performs comparably to CLVA on the GPU. 
Most importantly, it is $\sim$15x faster than CLVA on resource-constrained devices, taking just 0.39 seconds on the 6x i5-9500 CPU and 1.3 seconds on the Intel NUC device, as opposed to the respective durations of 5 seconds and 20.9 seconds for CLVA.
Moreover, we successfully demonstrate that our EdgeCLIPstyler model can load and run on a Raspberry Pi 3B+ device in 15 seconds. While not real-time, it is important to note that Raspberry Pi 3B+ is a very low-powered device and could not support loading the CLIPstyler, FastCLIPstyler and CLVA models due to memory issues. 
It should also be noted that though \cite{kwon2021clipstyler} presented a faster version of CLIPstyler, it was trained to support a single style at a time, making it incomparable to the other approaches. As a result, we have not included this model in our inference time benchmarking experiments.

The results demonstrate the high adaptability of our pipeline and its ability to achieve impressive performance even on resource-constrained devices. 
This opens up exciting possibilities for deploying our model in various settings, including low-power edge devices, embedded systems, and mobile applications. 

\subsection{Embedding space mapping}

Ghiasi \etal \cite{ghiasi_google} have successfully demonstrated that the embedding space of their style transfer network captures semantic information about styles. 
As we adopt their style embedding space to fit our text-style prediction network, we verify that our prediction network is also able to preserve the semantic information upon mapping from the text embeddings. 

To do so, we generate the text embeddings and corresponding style embeddings for various text prompts using our generalised text-style prediction network. 
Figure \ref{fig:t_sne_bmvc_general_prompts} illustrates the two-dimensional t-SNE plot of the style embeddings obtained by passing various combinations of generated text prompts through our text-style prediction network. 
As can be seen, our prediction network successfully maps semantically similar queries closer together in the dimensionally-reduced style embedding space.
More explorations on the embedding space visualisations can be seen in the supplementary section.

\section{Ablation Study}
\label{subsection:ablation_study}

\subsection{Effect of distribution loss}

Our proposed frameworks involve a text-style prediction network with a 100-dimensional output embedding space that is much larger than the valid region of Ghiasi network's embedding input. 
Hence, we propose the distribution loss to constrain the embedding towards the valid region of the input space of Ghiasi network, leading to much more visually pleasing results. 
This impact can be observed in \cref{fig:ablation_study} (a), where the stylised images have a more difficult time achieving content preservation and are heavily over-stylised.

\subsection{Effect of dataset generation using ChatGPT prompts}

The inclusion of ChatGPT-based prompts in our dataset generation step helps the model to generalise better and support a wide range of style queries without the requirement for a reference style image. We demonstrate the impact of ChatGPT-based prompts in \cref{fig:ablation_study} (b), where the inclusion of ChatGPT queries enhances the model performance, ensuring that the intricate details of the style prompt are captured better.

\section{Discussion and conclusion}
\label{subsection:conclusion}

By integrating pre-trained models from the purely vision-based style transfer domain into the CLIPstyler framework, we have developed the FastCLIPstyler model, capable of stylising a content image in a single forward pass. 
This model adeptly applies various styles and textures, as described in natural language, resulting in visually appealing images. 
We also introduce EdgeCLIPstyler, a model specifically designed for efficient LDAST execution on resource-constrained devices. 
As far as we know, this is the only model capable of effectively performing this task on low-power devices. 
Our experiments revealed that our models could generate stylised images free from undesirable artefacts often found in CLIPstyler outputs, all while operating several orders of magnitude faster. 
Additionally, our models outperform CLVA on numerous measurable metrics, including a human evaluation derived from a user survey.

The capability of our models, especially EdgeCLIPstyler, provides the foundation for a range of practical applications including real-time video conferencing background enhancements, sophisticated social media filters, and advanced photo editing tools.
Users can now execute local image editing directly on their mobile devices, eliminating the traditional dependency on remote server processing, and ensuring enhanced user experience along with stringent data privacy and security standards.

Despite these promising results, it is essential to address the limitations of our model to identify potential areas for improvement and future research. 
One notable limitation is the reliance on leveraging a pre-existing vision-based style transfer model with an explicit low-dimensional representation of the style. 
This constraint could be seen as restrictive, as recent research trends in the field have shifted towards models learning high dimensional implicit style representations by employing attention networks.

{\small
\bibliographystyle{ieee_fullname}
\bibliography{wacv}
}

\setcounter{section}{0}
\setcounter{figure}{0}
\setcounter{table}{0}
\setcounter{equation}{0}

\title{FastCLIPstyler: Optimisation-free Text-based Image Style Transfer Using Style Representations  (Supplementary Materials)}

\maketitle




\section{Model implementation details}

\subsection{Model architecture}

\begin{figure*}
\centering
\begin{tabular}{cc}
\includegraphics[height=7.25cm]{figs/architectureFastCLIPstylerFinalCameraReadyV3.pdf}&
\includegraphics[height=7.25cm]{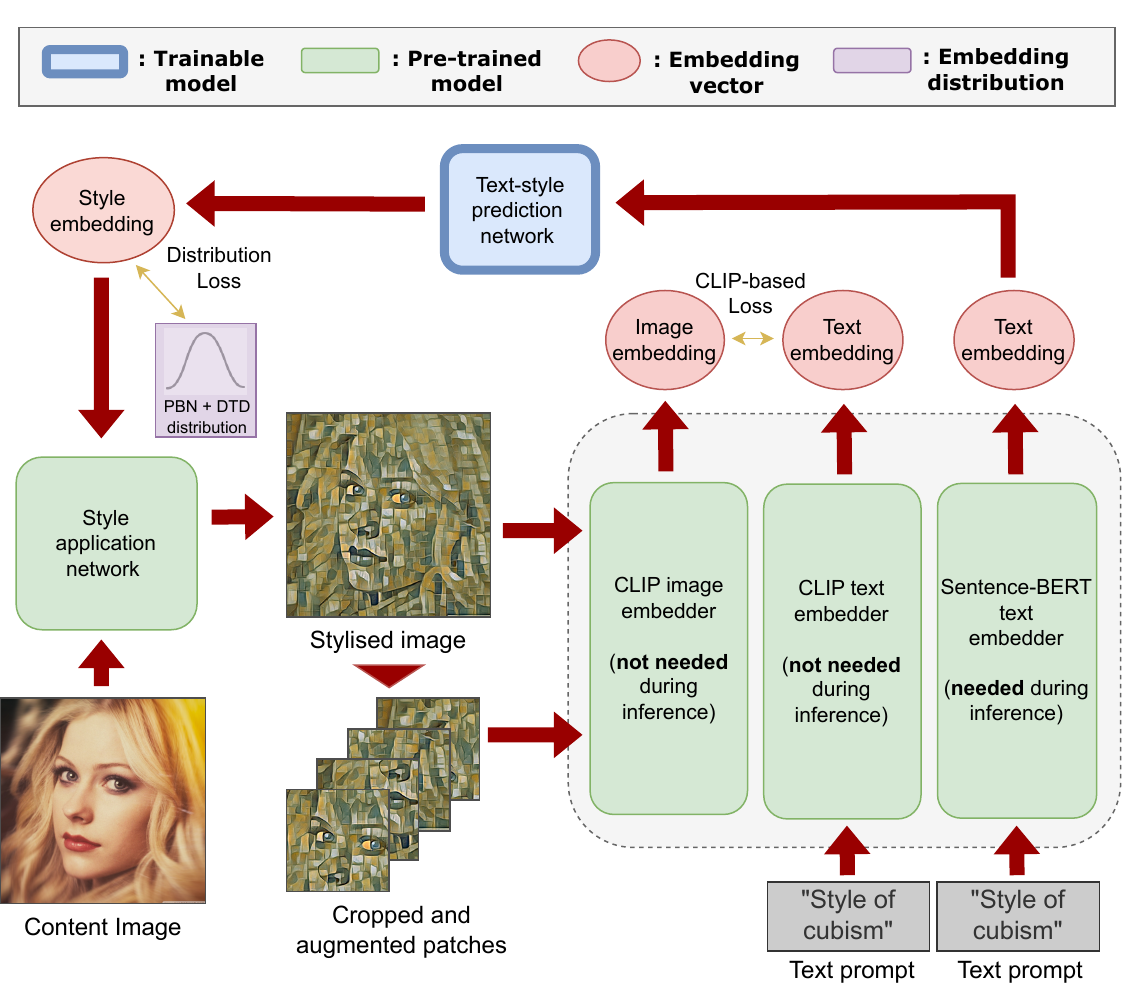}\\
(a) FastCLIPstyler & (b) EdgeCLIPstyler\\
\\
\end{tabular}
\caption{Architectural diagram of our two proposed approaches. 
}
\label{fig:architecture_fastclipstyler_edgeclipstyler}
\end{figure*}

In our research, we propose two main text-based image style transfer frameworks, namely FastCLIPstyler and EdgeCLIPstyler. 
While FastCLIPstyler performs stylisation in a single forward pass with state-of-the-art stylisation quality, EdgeCLIPstyler supports stylisation on edge devices with a slight trade-off in stylisation quality compared to FastCLIPstyler.
The architecture diagrams of FastCLIPstlyer and EdgeCLIPstyler can be seen in \cref{fig:architecture_fastclipstyler_edgeclipstyler}(a) and \cref{fig:architecture_fastclipstyler_edgeclipstyler} (b), respectively.

During the dataset generation phase, FastCLIPstyler and EdgeCLIPstyler optimise their respective text-style prediction networks for each prompt, to obtain the corresponding style embedding. To achieve this, both frameworks use the CLIP text and image embedder to compute the CLIP loss for optimisation. 
In order to generate the input text embedding for the text-style prediction network, FastCLIPstyler uses the CLIP text embedder, while the EdgeCLIPstyler framework uses the Sentence-BERT text embedder  \cite{reimers-2019-sentence-bert}.
During inference, EdgeCLIPstyler replaces FastCLIPstyler's CLIP text embedder with Sentence-BERT, eliminating the need for CLIP during inference and resulting in faster speed and compatibility with edge devices. It is to be noted that the CLIP image embedder is only required for loss computation, and is not utilised during inference for either framework.

\subsection{Text-style prediction network}

The text-style prediction network is a simple fully-connected feed-forward network that takes a text embedding as input and converts it into a 100-dimensional style embedding.

For FastCLIPstyler, a 512-dimensional text embedding is extracted from CLIP and passed as input to the text-style prediction network. The text-style prediction network comprises four hidden layers of 256, 256, 150, and 150 nodes, each with the Leaky ReLU \cite{maas2013rectifier} activation function with a negative slope of 0.2 as the activation function.
The final layer uses the tanh activation function to normalise the style representations between -1 and 1 since this is the range of values of style representation in the Ghiasi model \cite{ghiasi_google}.

On the other hand, EdgeCLIPstyler adopts the Sentence-BERT paraphrase-albert-v2 text embedder \cite{reimers-2019-sentence-bert} to extract a 768-dimensional text embedding, which is passed as input to the text-style prediction network. 
In order to accommodate the larger 768-sized text embedding, we have made modifications to the text-style prediction network to accept 768-dimensional as input.
The text-style prediction network, in this case, consists of five hidden layers with 512, 256, 256, 150, and 150 nodes, respectively, while adopting the same activation function setup as FastCLIPstyler.

\subsection{Model training and hyperparameters}

\subsubsection{Dataset generation}

A key step in the building of the FastCLIPstyler and EdgeCLIPstlyer models is the training of the text-style prediction network.
In order to train this network, we start by fitting the network for each text embedding individually and extracting the corresponding style embedding. 
This process is described by the following equation:
\begin{equation}
    e^{style}_i = f(e^{text}_i, \theta_i),
    \label{eq:data_gen}
\end{equation}
where $e^{style}_i$ is the style embedding, $f(\cdot)$ the text-style prediction network, $\theta_i$ is a parameter fitted for each prompt's CLIP text embedding $e^{text}_i$. 

\subsubsection{Generalised model training}

These style embeddings, obtained from the dataset generation step, are then used as `labels' for the model training stage, where the text-style prediction network is trained to map from the text embedding to its corresponding `label' style embedding using the standard mean-squared error (MSE) loss given by:
\begin{equation}
    MSE = \frac{1}{N} \sum_{i=1}^{N} \|e^{style}_i - \hat{e}^{style}_i\|^2,
\end{equation}
where $N$ is the number of text prompts in the dataset, $\hat{e}^{style}_i$ is the predicted style embedding at any particular stage in the training procedure, and $e^{style}_i$ is the pre-computed style embedding from Eq. \ref{eq:data_gen}.

The various hyper-parameters used to train the models and the time taken are enlisted in \cref{tab:benchmarking}. 

\begin{table*}[ht]
\centering
\begin{tabular}{ccccccc}
\hline
  & Learning & Epochs  & $\lambda_{dir}$ & $\lambda_{patch}$ & $\lambda_{dis}$ & Time taken\\ 
  & rate     &         &                 &                   &        &         \\ \hline
Data generation & $1 \times 10^{-3}$ & 150 & 500 & 1500 & 0.5 &  14.5 hours  \\ 
Model training & $1 \times 10^{-5}$ & 200 & -   & -    & -    & 3 minutes \\ \hline
\end{tabular}
\caption{Various hyperparameters used to train the different experiment runs and the time taken.}
\label{tab:benchmarking}
\end{table*}

\section{Datasets}
\subsection{Keyword-combination based prompts}
\label{subsect: prompts}

\begin{table*}[ht]
\centering
\begin{tabular}{ll}
\hline

 Categories & Keywords \\ \hline

\multirow{6}{*}{Colors} & White, Black, Green, Yellow, Pink, Purple, Blue, Orange, Red, \\
& Violet, Silver, Gray, Maroon, Teal, Aqua, Beige, Brown, Crimson,  \\ 
& Cyan, Gold, Greenyellow, Hotpink, Khaki, Magenta, Turquoise,  \\
& Fuchsia, Salmon, Seashell, Chocolate, Peru, Whitesmoke, Honeydew, \\
& Rosybrown, Saddlebrown, Seagreen, Slategray, Steelblue, Indianred, \\
& Olive, Lime, Navy, Lavender, Indigo, Ivory   \\
 \hline

\multirow{5}{*}{Objects} & Wool, Color pencil, Pencil, Brush, Color brush, Crystals, color, \\
& Crystals, Color painting, Crayon, Lines, Watercolor, Stone wall,  \\ 
& Cloud, Underwater, Fire, Metal, Lightning, Wave, Flames, Leafy,   \\
& Grassy, Darkness, Wooden, Snow, Iceberg, Cartoon, Comic, Squares,  \\
& Lava, Vines, Magma, Desert sand, Water waves, Lace \\
 \hline

Art styles & Acrylic, Oil Painting, Mosaic, Cubism, Monet \\ \hline

\multirow{3}{*}{Textures} & Crystalline, Cracked, Crosshatched, Fibrous, Freckled, Grid, \\ 
& Honeycombed, Meshed, Perforated, Porous, Scaly, Smeared, Studded, \\ 
& Swirly, Veined, Waffled, Woven, Wrinkled, Pleated, Sprinkled, Knitted \\ \hline

\end{tabular}
\caption{Keywords used for forming the combination prompts dataset}
\label{tab:keywords_general_prompts_dataset}
\end{table*}

\begin{table*}[ht]
\centering
\begin{tabular}{ll}
\hline

 Categories & Prompts \\ \hline

\multirow{3}{*}{Keyword combination prompts} & White color, Blue color pencil, Desert sand, \\
&  Style of maroon wool, Orange cracked, Aqua colour style\\ 
& Mosaic style painting, Cracked acrylic painting, etc. \\ \hline

\multirow{4}{*}{ChatGPT generated prompts} & Ocean waves and serene calmness, \\ 
& Glittering stars in the night sky, \\
& A dreamlike landscape bathed in the soft light \\
& evoking feelings of tranquillity \\
\hline
 
\end{tabular}
\caption{Samples of keyword combination prompts and ChatGPT generated prompts.}
\label{tab:prompts_dataset_examples}
\end{table*}

To train our model, we constructed a dataset of queries by combining a list of keywords that included 44 different shades of colours, 21 textures, 5 art styles, and descriptions of 34 real-world objects with distinguishable textures. The list of keywords can be seen in \cref{tab:keywords_general_prompts_dataset}.
Various random combinations such as colour-texture, colour-art, colour-object, art-texture, and art-object were created to generate the dataset of 4,302 prompts. Examples of the keyword combination prompts can be seen in \cref{tab:prompts_dataset_examples}.
We followed specific strategies to preserve the legibility of this final list of generated prompts.
For instance, we appended the word `colour' at the end of each of the 44 shades.
This led to the removal of ambiguity for colours such as `salmon', `seashell', `chocolate', etc. 
In the case of textures, we selected 21 different textures from the DTD dataset by filtering out textures that seemed visually repetitive.
We randomly combined these textures with the set of colours and text descriptions of real-world objects such as `stone wall', `cloud', `fire' to name a few.
Some examples of such combinations include `maroon crosshatched', `yellow freckled', `black wrinkled', etc.
Additionally, the dataset consists of five different art styles, namely `acrylic', `oil painting', `mosaic', `cubism', and `monet', which too were randomly combined with the set of colours and textures.
The resulting combinations include prompts such as `pink oil painting', `mosaic cloud', and `studded cubism', among others.
Furthermore, we randomly augmented the final list of prompts with leading phrases such as `style of', `a pixelated photo of', `a blurry photo of', `a sketch of a', and `a black and white photo', as well as with trailing phrases such as `style' and `style painting'.
Examples after this augmentation include `style of blue colour', `a blurry photo of white wool', `a pixelated photo of cyan stone wall', and many more.
Experimental results show that our network generalises while being trained on this dataset and can generate stylised images with unseen queries of quality comparable to the current state-of-the-art.

\subsection{ChatGPT based prompts}

CLVA's \cite{fu2022language} research effectively illustrated the potential of the ArtEmis dataset \cite{artemis} for enhancing the training of style transfer systems.
While integrating the ArtEmis dataset into our style transfer model training, we identified a mismatch between the dataset's content-descriptive prompts and our requirement for style-focused prompts. 
ArtEmis prompts, though rich and varied, contains a large number of annotations that refer to visible contents rather than capturing the essence of artistic style.
To rectify this, we engaged ChatGPT \cite{brown2020language} to adapt and generate prompts that were more aligned with our model's objectives. 
We selected representative prompts from ArtEmis that subtly suggested style and provided ChatGPT with guidelines to craft prompts that foreground elements of style such as texture, colour scheme, and compositional flow.
The utilisation of ChatGPT for this task resulted in an enriched set of 1,500 prompts that more closely cater to the model's need for stylistic abstraction. 
These prompts avoid direct content representation, fostering a training environment that emphasises the stylistic interpretation rather than the literal depiction. 
The effectiveness of this tailored dataset is evident in the improved model performance and is exemplified by the curated examples in \cref{tab:prompts_dataset_examples}, which demonstrate the prompts' adherence to the model's style-centric learning paradigm.

\section{Embedding space mapping}
\label{subsect: embedding space}

\begin{figure*}
    \begin{tabular}{cc}
      \includegraphics[width=0.45\textwidth]{figs/t_sne_bmvc_general_prompts.pdf} &
      \includegraphics[width=0.45\textwidth]{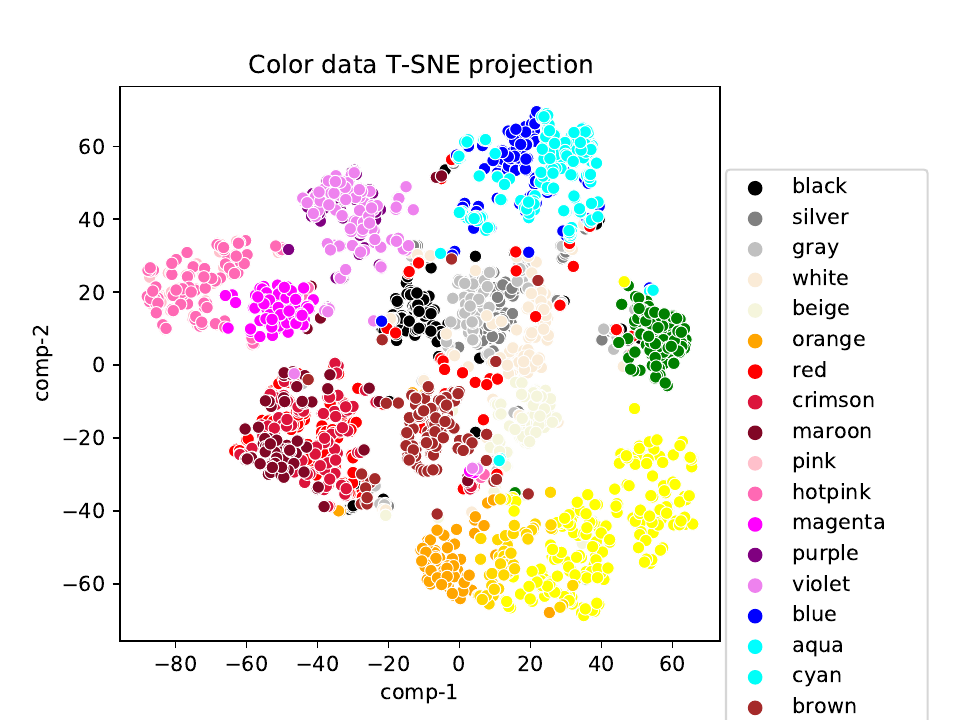} \\
     (a) General objects t-SNE visualisation & (b) Colors t-SNE visualisation \\[6pt]
      \includegraphics[width=0.45\textwidth]{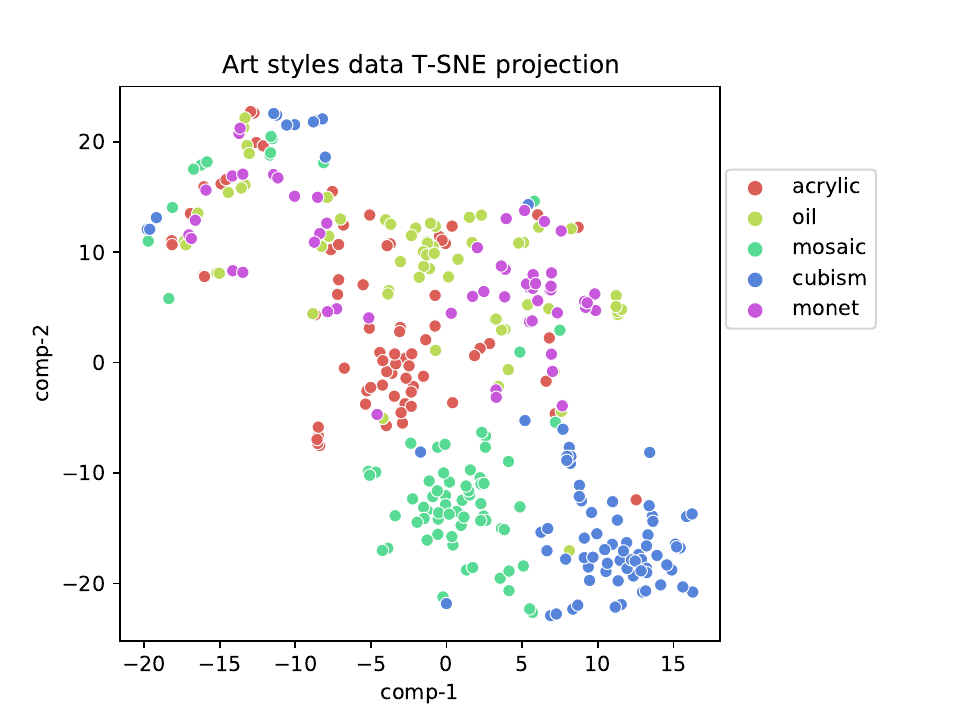} &   \includegraphics[width=0.45\textwidth]{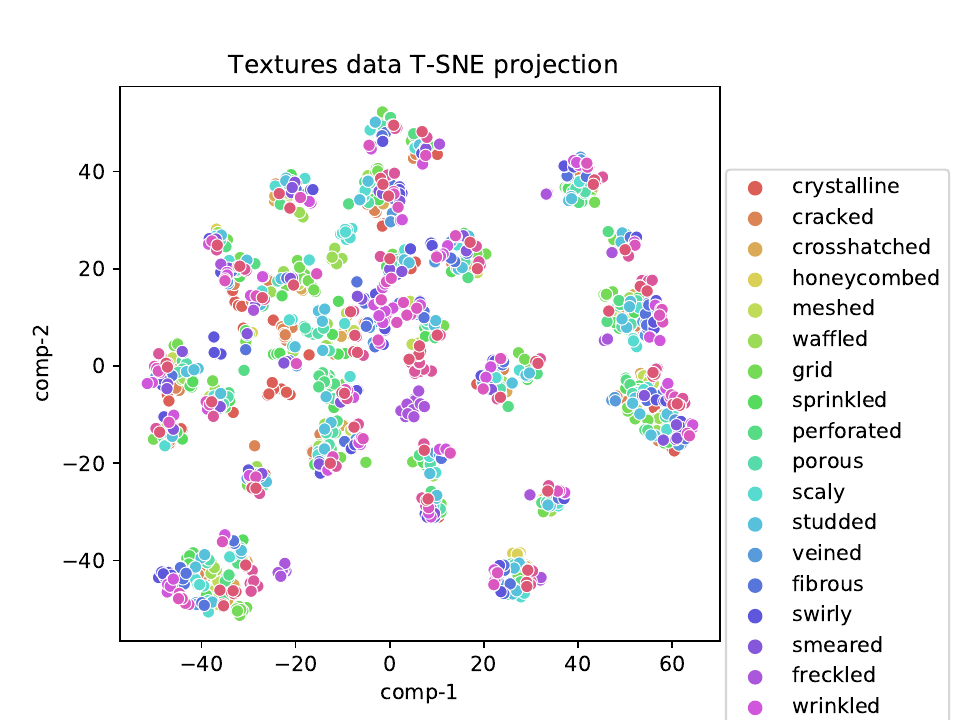} \\
    (c) Art styles t-SNE visualisation & (d) Textures t-SNE visualisation \\[6pt]
    \end{tabular}
    \caption{The t-SNE visualisation of style embedding space upon mapping from the text prompts.}
    \label{fig:t_sne_visualisation}
\end{figure*}

Ghiasi \etal \cite{ghiasi_google} have successfully demonstrated that the embedding space of their style transfer network captures semantic information about styles. 
As we adopt their style embedding space to fit our text-style prediction network, we verify that our prediction network is able to preserve the semantic information upon mapping from the text embeddings. 

To do so, we generate the text embeddings and corresponding style embeddings for various text prompts using our generalised text-style prediction network. 
As the style embedding vectors are 100-dimensional, for visualisation purposes, we perform dimensionality reduction using t-SNE \cite{vandermaaten2008tsne}.
\Cref{fig:t_sne_visualisation} illustrates the two-dimensional t-SNE plot of the style embeddings obtained by passing various combinations of generated text prompts through our text-style prediction network. As can be seen, our prediction network successfully maps semantically similar queries closer together in the dimensionally-reduced style embedding space. 

For \cref{fig:t_sne_visualisation}(a), 
we built a dataset of 1,516 text prompts comprising 15 keywords \{fire, flames, magma, lava, leafy, vines, grassy, wave, water waves, underwater, cartoon, comic, lightning, cloud, darkness\} combined with various colours and textures, forming prompts like `a blurry photo of red underwater', `cracked leafy style', and `a pixelated photo of magma'.
We also grouped these keywords together into semantically meaningful sets, including \{fire, flames, magma, lava\}, \{leafy, vines, grassy\}, \{wave, water waves, underwater\}, \{cartoon, comic\} and \{lightning, cloud, darkness\}.
We demonstrate that for prompts having similar semantic meanings, their style embeddings lie very close together in the embedding space. For example, prompts containing `fire' and `flames' or `magma' and `lava' almost overlap in the embedding space. 
We see similar behaviour for all the above-mentioned semantically meaningful sets.
Due to the complexity of these prompts formed from pairing keywords with colours and textures, we witness some outliers lying outside the main distribution. However, in most cases, they preserve a similar structure of semantically meaningful queries.

For the visualisation of colour-based prompts (\cref{fig:t_sne_visualisation}(b)), we generated 2,968 prompts containing 22 colours combined with various general objects, art styles, and textures. 
We can see similar colour-based prompts lie closer in the embedding space.
As for art style-based prompts (\cref{fig:t_sne_visualisation}(c)), we generate 209 prompts for five art styles \{acrylic, monet, oil, cubism, and mosaic\}, combined with various colours and textures. We can see art styles such as `cubism' and `monet', which have a similar style, lying close in the embedding space. Additionally, `acrylic', `oil', and `monet', again with similar styles, lie close in the embedding space.
We also visualise 1,516 texture-based prompts (\cref{fig:t_sne_visualisation}(d)), comprising 21 textures combined with various colours and objects. While the plot is scattered due to combining with various other well-preserved colours and objects, it seems to preserve semantic meaning within these scattered distributions.

\section{General model performance}
\begin{figure*}
    \centering
    \includegraphics[width=0.75\textwidth]{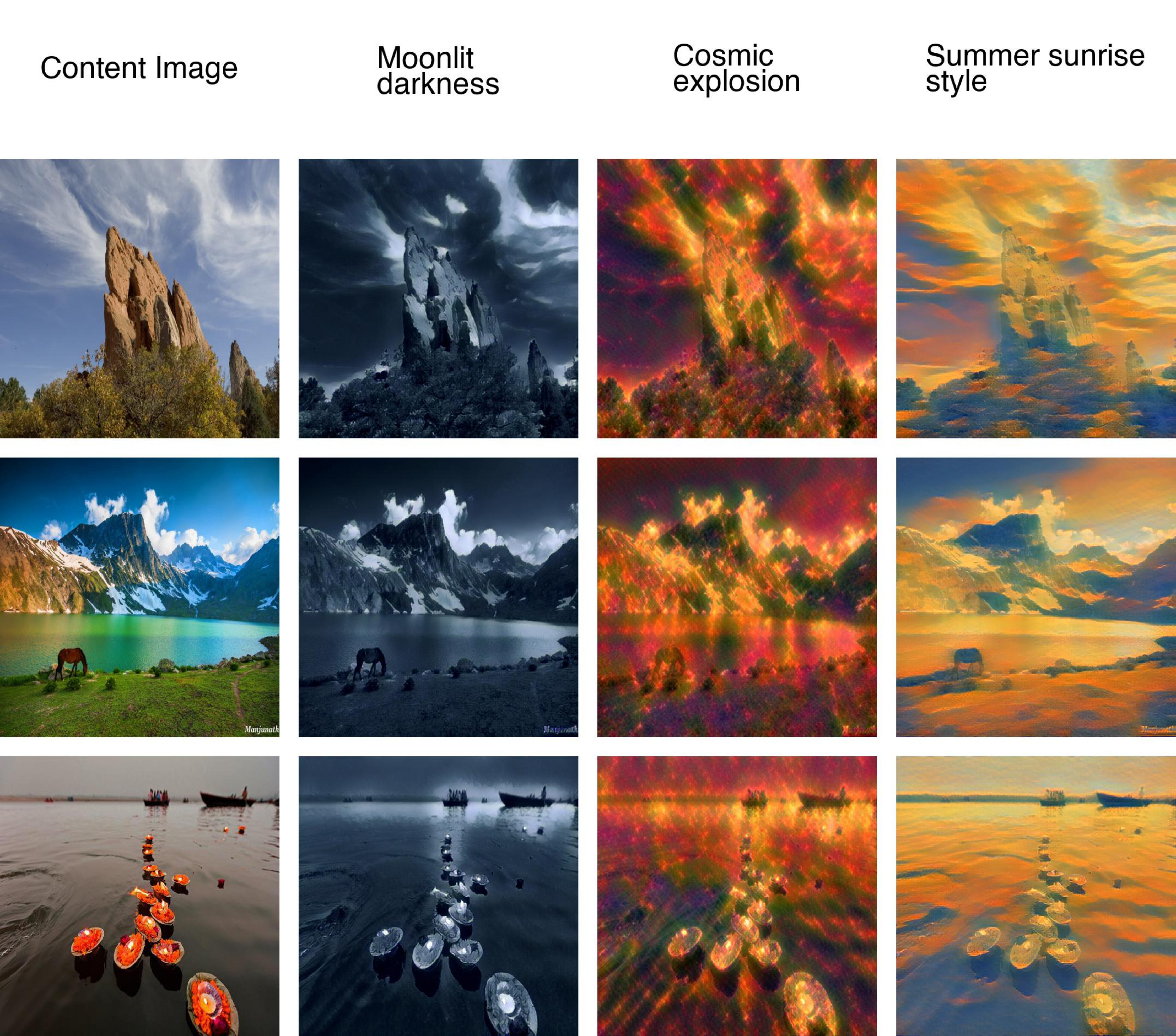}
    \caption{Performance of our FastCLIPstyler model on general style prompts.}
    \label{fig:general_model_performance_general}
\end{figure*}

\begin{figure*}
    \centering
    \includegraphics[width=\textwidth]{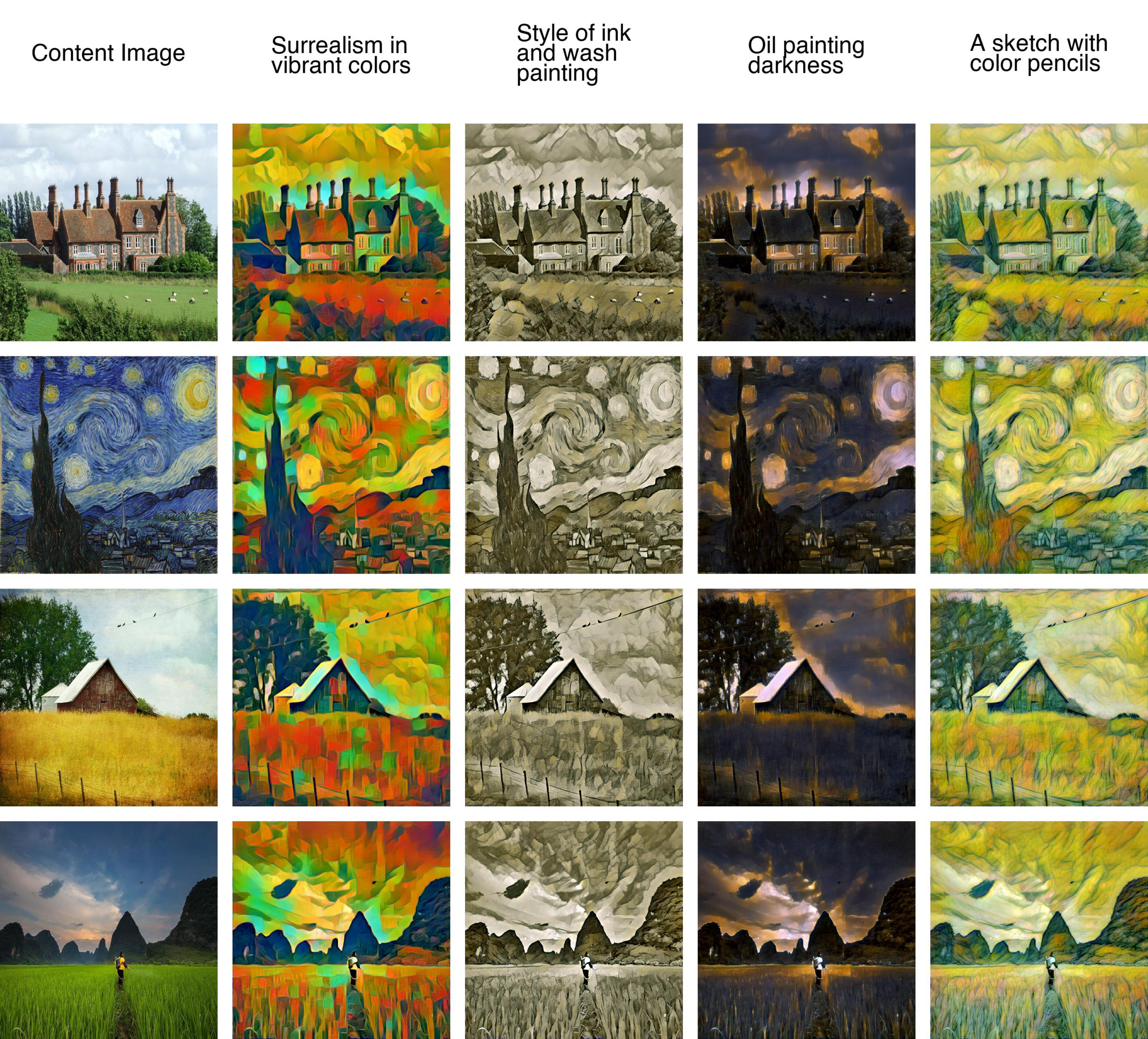}
    \caption{Performance of our FastCLIPstyler model on art style prompts.}
    \label{fig:general_model_performance_art_styles}
\end{figure*}

\begin{figure*}
    \centering
    \includegraphics[width=\textwidth]{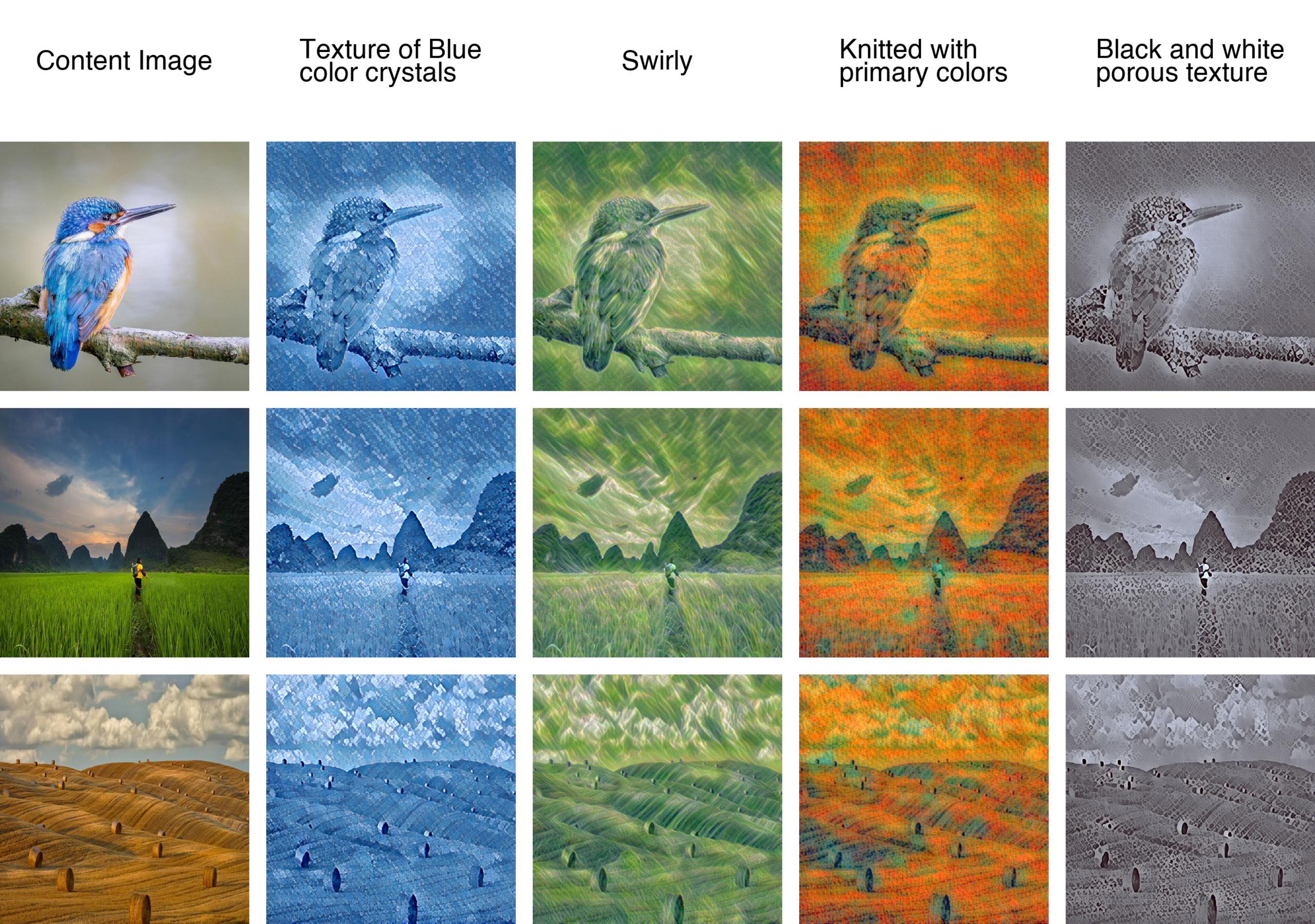}
    \caption{Performance of our FastCLIPstyler model on texture style prompts.}
    \label{fig:general_model_performance_texture}
\end{figure*}

\begin{figure*}
    \centering
    \includegraphics[width=\textwidth]{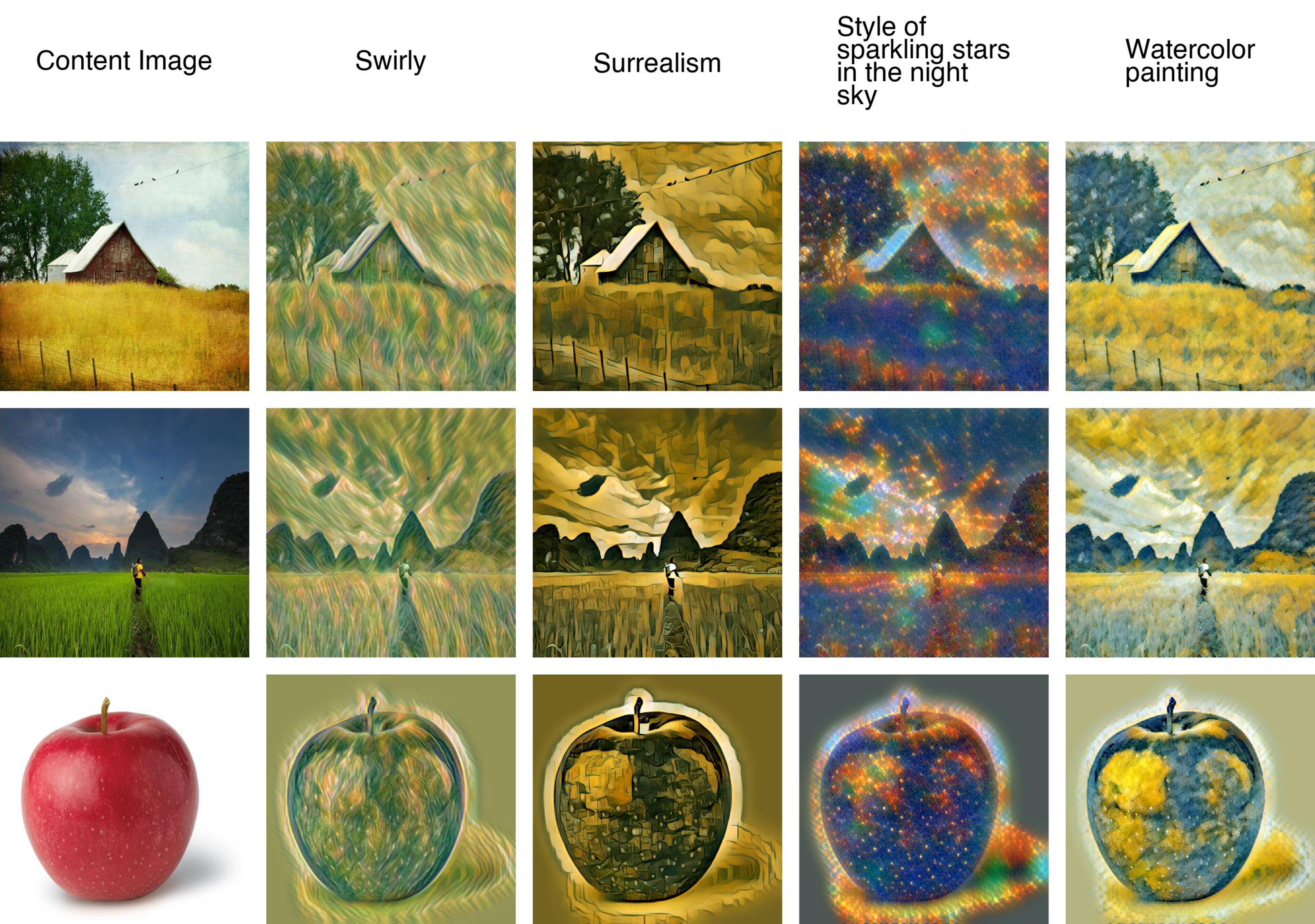}
    \caption{Performance of our EdgeCLIPstyler model on various prompts.}
    \label{fig:general_model_performance_edge}
\end{figure*}

We share additional stylisation results of our models on various general prompts, art styles, and textures. The results for FastCLIPstyler can be seen in \cref{fig:general_model_performance_general}, \cref{fig:general_model_performance_art_styles}, and \cref{fig:general_model_performance_texture}.
The results for EdgeCLIPstyler can be seen in \cref{fig:general_model_performance_edge}.

\section{Quantitative evaluation - Human evaluation}

\begin{figure*}
    \centering
    \includegraphics[width=\linewidth]{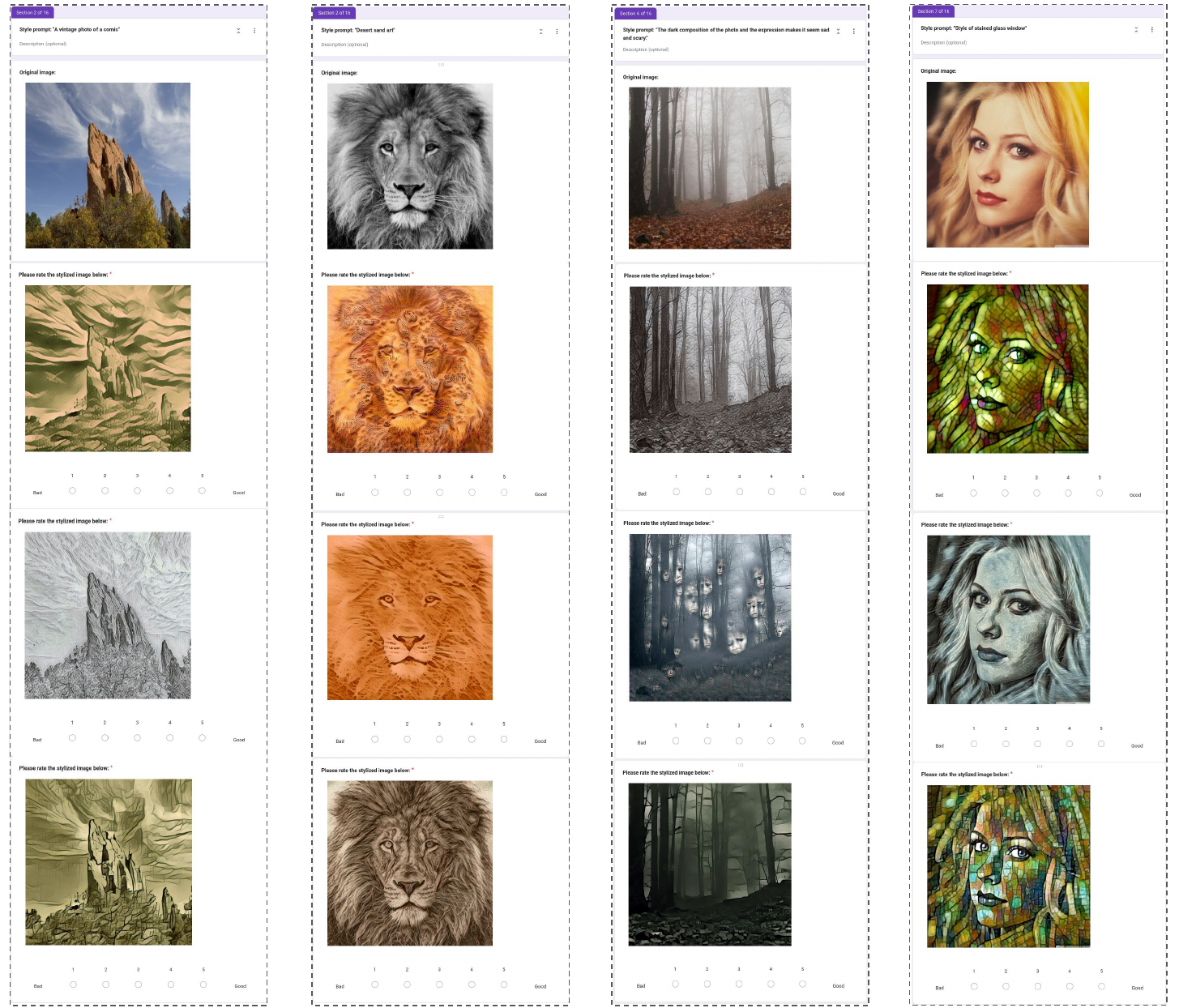}
    \caption{Sample screenshots of the forms sent to the participants.}
    \label{fig:form_sample}
\end{figure*}

Human evaluation was done by means of an online form survey. 
We created four distinct forms, each with 20 questions.
A total of 75 participants were randomly given one of these forms.
All forms start with giving information on how style transfer works and what the desirable and undesirable properties of a stylised image are with visual examples.
Sample images of the form are shown in \cref{fig:form_sample}.
Each section of the form shows a style prompt and displays a content image and its corresponding stylised images generated by different models.
The participants were then asked to rate from 1 to 5 the quality of each image.
The final scores for each model are then calculated as the mean of the scores from the form responses.

The prompts and associated content images were chosen to ensure a fair representation. 
We randomly select prompts from various databases, such as ArtEmis and DTD, which were utilised for model evaluation in the original research by CLVA.
Similar methodologies were employed for the selection of prompts from CLIPstyler and our own model evaluations. 
This strategy ensures an unbiased comparison across all models under review.
For fairness purposes, the stylised images shown to the participants were not labelled with their corresponding models and were arranged randomly to reduce potential order bias.

\begin{figure*}
    \centering
    \includegraphics[width=\linewidth]{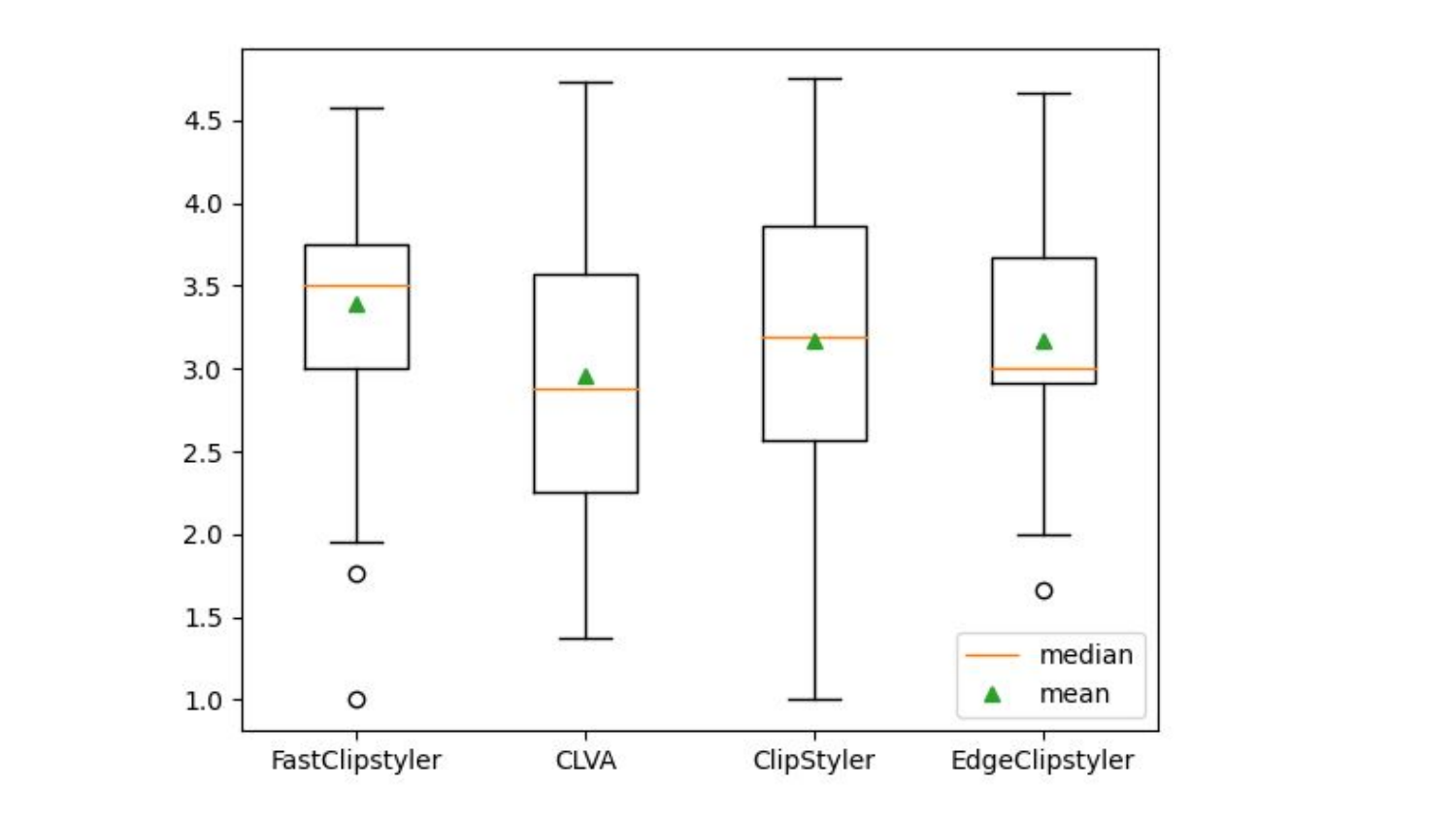}
    \caption{Result from user study as box plot}
    \label{fig:user_study_boxplot}
\end{figure*}

For text prompt selection, we randomly selected a subset of text prompts generated for training the text-style prediction network. 
We additionally create a set of general prompts for human evaluation.
For content images, we randomly selected different content images from our qualitative evaluation experiments. Then from a pool of final prompts, we split them into four evaluation forms, 20 prompts each. Additional user study result is shown in \cref{fig:user_study_boxplot} as a box plot of rated scores from 1 to 5 for each of the method.

\begin{figure*}
    \centering
    \includegraphics[width=\linewidth]{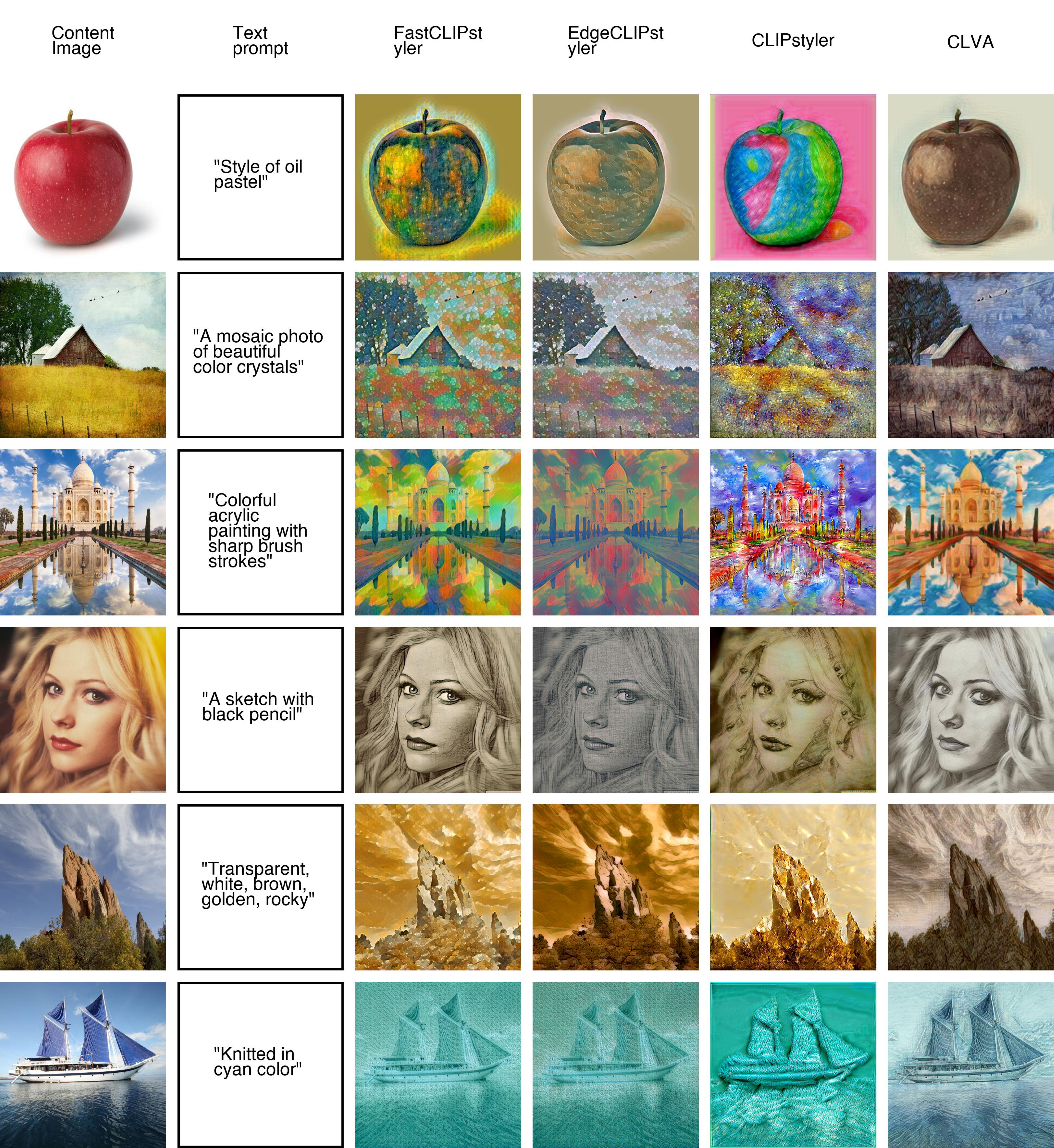}
    \caption{Comparison of our models with CLIPstyler and CLVA.}
    \label{fig:content_samples}
\end{figure*}

\section{Additional comparison results}

We compare the performance of our model to the state-of-the-art text-based image style transfer approaches, namely CLIPstyler and CLVA. 
Experimental results show that our FastCLIPstyler results are on par with the quality of these other approaches in general.
While CLIPstyler produces impressive stylisation results due to the advantage of optimising the process at run-time, it tends to sometimes over-stylise images and introduce artefacts as well.
Over-stylisation in this context refers to the case when the style is applied too heavily, to the point where it overwhelms the content of the image or even adds artefacts (unwanted objects that represent content instead of style) onto the image.
One reason for this could arise from the high flexibility in the architecture of CLIPstyler. 
The CNN that styles the image does not have a direct understanding of artistic paintings or specific art styles. 
Instead, it achieves this indirectly through the loss function defined by CLIP. 
As a result, the model has `too much freedom' in editing the image, which can sometimes lead to undesired results.

Although CLVA can stylise images in a single forward pass, it falls short in terms of generalising to diverse prompts.
Our FastCLIPstyler model provides stylisation results on par with these approaches and generalises to diverse prompts while being able to produce results in a single forward pass. 
Our EdgeCLIPstyler supports edge devices with a minor drop in performance.
A qualitative comparison of our models with other recent state-of-the-art text-based image style transfer approaches is shown in \cref{fig:content_samples}. 

\section{Negative social impact}
Text-based image style transfer has made it easy to manipulate an image with a style description. 
This is both beneficial and also can have negative social impacts. 
Since a reference style image is not required, users are able to manipulate any image using just words.
This might have the highest negative impact on people who work in the graphical design area. 
If developed further, it could potentially replace those jobs in some cases that do not require sophisticated work. 
Also, we do not know the full capability of this model. It might be able to transform an image in unexpected ways that might cause harm to the user.


\end{document}